\documentclass[review,3p]{elsarticle}

\makeatletter
\def\ps@pprintTitle{%
 \let\@oddhead\@empty
 \let\@evenhead\@empty
 \let\@evenfoot\@oddfoot}
\makeatother

\usepackage{rotating}

\usepackage{threeparttable}

\usepackage[english]{babel}
\usepackage[utf8]{inputenc}
\usepackage{algorithm}
\usepackage[noend]{algpseudocode}

\usepackage{lineno}

\usepackage{graphicx}
\usepackage{amsmath}
\usepackage{url}
\usepackage[bookmarks=true%
,bookmarksnumbered=true%
,hypertexnames=false%
,colorlinks=true%
,linkcolor=blue%
,citecolor=blue%
,urlcolor=blue%
]{hyperref}

\usepackage{subfigure}

\usepackage{changepage}
\usepackage{booktabs}

\usepackage{multirow}
\usepackage{verbatim}
\usepackage{breqn}
\usepackage{enumitem}  

\usepackage{amsthm}
\usepackage{algorithm}

\usepackage{relsize}

\biboptions{authoryear}

\title{Improved two-stage hate speech classification for twitter \\ based on Deep Neural Networks}

\usepackage{algpseudocode}

\begin{document}

\hypersetup{
 breaklinks=true   
}

\author[rvt]{Georgios K. Pitsilis\corref{corresp}}
\ead{georgios.pitsilis@gmail.com}
\address[rvt]{Independent scholar}
\address[]{Computer Science Research \& Development, \\Athens, Greece\\}

\cortext[corresp]{Corresponding author}

\begin{abstract}

	Hate speech is a form of online harassment that involves the use of abusive language, and it is commonly seen in social media posts.
	This sort of harassment mainly focuses on specific group characteristics such as religion, gender, ethnicity, etc and it has both societal and economic consequences nowadays. 
	The automatic detection of abusive language in text postings has always been a difficult task, but it is lately receiving much interest from the scientific community.
	This paper addresses the important problem of discerning hateful content in social media.
	The model we propose in this work is an extension of an existing approach based on LSTM neural network architectures, which we appropriately enhanced and fine-tuned to detect certain forms of hatred language, such as \emph{racism} or \emph{sexism}, in a short text.
	The most significant enhancement is the conversion to a two-stage scheme consisting of \emph{Recurrent Neural Network (RNN)} classifiers.
	The output of all \emph{One-vs-Rest (OvR)} classifiers from the first stage are combined and used to train the second stage classifier, which finally determines the type of harassment.
	Our study includes a performance comparison of several proposed alternative methods for the second stage evaluated on a public corpus of 16k tweets, followed by a generalization study on another dataset.
	The reported results show the superior classification quality of the proposed scheme in the task of hate speech detection as compared to the current state-of-the-art.

\end{abstract}

\begin{keyword}
text classification
\sep micro-blogging
\sep hate speech
\sep recurrent neural networks
\sep Twitter
\sep One-vs-Rest classification
\end{keyword}


\maketitle


\section{Introduction}

The global popularity of social media platforms, where users may express their opinions and interact with others online, has resulted in an explosion in the volume of data produced. 
Social media may reflect public sentiment and strongly influence people's opinions on various events. 
However, social media platforms are not always safe, and they can become vulnerable to the spread of aggressive content expressed in the form of abusive language. 
That may have implications for the users' online experience and the community in general.

\emph{Cyber-hate} or \emph{cyber-violence}, \cite{Miro2016}, refers to any form of violent communication over the internet, and terms like \emph{toxic comments} or \emph{hate speech} are used to describe any abusive language used on social media. 
Unlike \emph{cyber-violence}, \emph{hate speech} may affect entire communities as well as individuals, \cite{Blaya2018}.

Many social media service providers, including Facebook and Twitter, have expressed serious concerns over the presence of offensive text in posts.
Despite their efforts to implement tools for users to report abusive content, they have been criticized for not doing enough to prevent the use of toxic language.
Human moderators is one of the current options being used to address the problem, but it has serious limitations, including the substantial effort required, the high operational cost, the lack of scalability, and the significant influence on the system's response time. 
Various automatic detection techniques have been developed in recent years to address this issue and deal with abusive content, \cite{hosseini2017deceiving,twohat:2020,Hatebase:2020}.
The most recent tactic implemented by service providers is the termination of any reported misbehaving users' accounts.

The scientific community has responded to the toxic language problem with many suggested solutions, \cite{AlHassan2019}.
Numerous techniques rely on Natural Language Processing (NLP) solutions that use Part-Of-Speech (POS) rules to address the problem. 
Nonetheless, there has recently been a trend toward using Machine Learning (ML) solutions. 
The high complexity of the NLP techniques is the main reason for their declining popularity, compounded by the fact that computing power in recent years is becoming increasingly cheaper and easier to use for training ML algorithms and supervised learning models in general.
Modern ML approaches, such as Neural Networks and Deep Learning (DL) architectures, are gaining popularity in many natural language processing tasks, such as question answering, machine translation, and other text classification tasks, due to their ability to achieve good performance when trained on large sets of labelled data. 

Automatic moderation tools, whether based on POS techniques or entirely on Artificial Intelligence (AI), usually fail to classify specific forms of hateful content appropriately.
\emph{Sarcasm}, \emph{sexism} and \emph{racism} are among them, with the main reason being that offensive context in textual posts is often phrased using out-of-vocabulary words or expressed indirectly via regular non-toxic words.
In fact, POS-based solutions fall short in this respect since they are primarily reliant on the language used in the text. 
For a similar reason, many of the existing supervised learning algorithms proposed for the \emph{hate speech} problem, including state-of-the-art Neural Network solutions, are deemed inadequate in classifying text that contains out-of-vocabulary words.
The main cause of this is their high dependency on word embeddings like \emph{GloVe} by \cite{GloVe}, \emph{Word2Vec} by \cite{word2vec}, or \emph{FastText} word representations by \cite{Bojanowski2017}, which used as pre-trained vectors in the encoding of input text.

The concept behind using word embeddings in the mechanism is to represent words as vectors, with semantically similar words assigned to vectors close to each other in the vector space. 
Nevertheless, any \emph{hate speech} written in a way that any offensive terms are hidden behind slang words unrelated to any existing word embeddings, will remain undetected.
\cite{mishra-etal-2018-neural} for example reports a case of a data set of tweets containing 16 thousand unique tokens, 5.5 thousand of which are not in the English dictionary. 
Among these unknown tokens, 600 are in \emph{racist} tweets, while 1.6 thousand are in \emph{sexist} tweets.
According to \cite{seganti-etal-2019-nlpr}, FastText 1M embeddings only contain around half of the words in the corpus of the typical \emph{hate speech} dataset used in the SemEval-2019 competition task. 

Bidirectional Encoder Representations from Transformers (BERT) models by \cite{Devlin2019BERTPO}, provide pre-trained vector representations of words that capture their contextual relationships in the sentence.
Despite breaking multiple records in language-related tasks, BERT falls into the same category as architectural solutions that rely heavily on pre-trained vectors of input text, and so it suffers from the same flaws.

Incorporating user-specific features, like someone’s history of \emph{hate speech} postings, into the decision-making mechanism for assessing the degree of hatefulness that may exist in his/her future postings has been shown to provide a significant performance benefit, \cite{qian-etal-2018,fehn-unsvag-gamback-2018,Pitsilis2018,dadvar2013,mishra-etal-2018}.
However, this technique has four main drawbacks:
First, it suffers from systemic bias in discriminating against users who tend to use hatred language, thus posing ethical concerns regarding the freedom of expression.
Second, it requires disclosing user identity information to third parties (e.g. service providers).
That is because users' posting history contains information useful for feature extraction throughout the training process.
Third, any reliance on user history data may prove ineffective when applied to new users who do not have a previous record or to others who intentionally change their behaviour in order to get misclassified.
Fourth, the metadata describing a users’ history may vary from one service to another, limiting the overall applicability of a system.
It is worth mentioning the reduced applicability of this solution only to a small number of systems that still include user profiling in the process.
We should also highlight that, due to concerns about anonymity, most datasets appropriate for training a \emph{hate speech} classifier lack such user-related information, making profile-based categorization impractical in reality. 

Also of great importance for the classification service is to provide comprehensive and  trustworthy output.
Producing detailed classification output, which indicates specific forms of toxicity such as \emph{racism}, \emph{sexism}, \emph{misogyny}, \emph{politics} and \emph{religion}, instead of a binary expression (e.g. toxic/non-toxic), could be a step forward.
The above forms of toxicity are not mutually exclusive, which means \emph{hate speech} detection is, in fact, a multi-label classification challenge.
With this in mind, a moderation system must be flexible enough to make decisions based on specific rules.
\\

Another crucial fact concerning \emph{hate speech} classification is the variation in prediction quality levels observed between each toxicity class.
More particular, studies reported by scientists in the field, \cite{Pitsilis2018}, indicate that \emph{racism} content is much harder to identify than \emph{sexism}, resulting in lower classification scores for that class, \cite{Arango:2019}. 
That is because, in many cases, \emph{racism} is conveyed without the use of toxic words, resulting in false negatives. 
For example, the non-toxic word \emph{"black"} is occasionally used for offending a specific group of people by concealing the offensive context behind a non-toxic phrase.
For the opposite reason, there may be false positives, e.g. when using a toxic word for self-referential comments like \emph{"I feel like an idiot now"}, which usually leads to incorrect classification.
In our opinion, some improved ability in detecting similar cases would be a serious step towards improving the classification performance.

Considering the above points, we summarize the design goals for \emph{hate speech} classification algorithms as the following:
\emph{i)} 
Language-agnostic classification with no dependency on pre-trained word embeddings.
\emph{ii)} 
Utilizing the textual content alone for feature extraction instead of being devised from user profile data.
\emph{iii)}
Preference for using contextualized features over contextualized word embeddings.
\\

In this paper, to address the above issue, we propose a two-stage classification scheme.
The first stage consists of multiple One-vs-Rest (OvR) Deep Learning classifiers, with their output being combined in the second stage to train another classification scheme. 
The second stage is where the class of abusive content is finally determined.

OvR is a heuristics method used for multi-class classification that employs multiple binary classifiers.
This is accomplished by splitting the training data into several binary sets, one for each class, a strategy that works well when dealing with a small number of classes.
Other researchers, \cite{Zhou2012} have noted that it is preferable for the learning process when there are numerous labels for expressing the sub-concepts of the original data (e.g. sub-classes of hatred language: sexism / racism / neutral) than when only binary-labelled data is available (e.g. aggressive / non-aggressive).
Furthermore, incorporating OvR classifiers in the model, rather than just employing a single multi-class classifier, is preferable for its flexibility for fine-grained classification while yet providing improved classification quality, \cite{ICWSM113841}.
Given the above, providing a solution capable of dealing with various toxicity classes is still feasible.
Although the employment of OvR schemes alone in text classification is not in itself a significant contribution, our aim in this study is to demonstrate how fine-grained classification for toxic language can be achieved by combining various conventional concepts into a single mechanism.
In addition, although our experimentation is limited to distinguishing only \emph{sexism} and \emph{racism} from the \emph{neutral} text, our concept can easily be extended to support more classes of toxicity and text content written in many languages other than English.

Our main contributions are:
\emph{i)} A neural network-based architecture for classifying \emph{hate speech} content, consisting of several OvR Deep Learning classifiers into an ensemble scheme.
\emph{ii)} A model for \emph{hate speech} detection with language-agnostic ability, which in theory can be trained on labelled data in the language of preference.
\emph{iii)} A comparative experimental evaluation of the proposed architecture on a dataset of tweets, in relation to other state-of-the-art solutions. 
\emph{iv)} A generalization study of the proposed model by testing on another set of \emph{hate speech} data with a comparison of performance against state-of-the-art algorithms.

The rest of the paper is organized as follows: 
In section \ref{statement}, we elaborate on the details of the problem of \emph{hate speech} that we come to address, and in section \ref{relWork}, we refer to existing work in the field. 
In section \ref{model}, we present our solution in more detail, and in section \ref{eval}, there is a comparative evaluation of its performance.
Finally, in section \ref{conclusion}, we summarize our contribution and outline possible future work.

\section{Problem statement - Motivation}
\label{statement}

The problem we address in this paper is described as follows: 
Given a set of text postings submitted by different users, each posting may contain several words.
Let "Sexism", "Racism", and "None" be three categories, with the first two stating the type of abuse in the textual content, while the third one stating the absence of abuse.
In addition, each posting is limited to being a member of only one class of hatred language.
Given that postings may contain out-of-vocabulary words, we assume that there is no other available information to associate postings and users who submit them.
Based on these facts, the challenge is determining the class to which a new unlabeled text posting belongs.

The research question addressed in this work is:

\begin{quote}\it
How to effectively determine the class of hatred language included in some postings that may contain out-of-vocabulary words while omitting any profile-related information that could make the posting users identifiable?
\end{quote}

To answer this question, our main goals are summarized as follows:

\begin{itemize}
    \item To develop a novel classification method that can improve the state-of-the-art, in terms of classification performance, without sacrificing users' privacy.
    \item To develop a method for fine-grained detection of abusive language independent of the linguistic rules or vocabularies used and investigate the impact of such requirements on its performance.
\end{itemize}

Note that, in many existing solutions, to improve performance, they resort to tactics of utilizing users’ posting activity data along with the text postings themselves.
Meanwhile, other non-profiling oriented solutions are only effective for certain classes of \emph{hate speech}.
In our opinion, the \emph{hate speech} detection tasks should be compliant with the privacy concerns of GDPR\footnote{General Data Protection Regulation https://eugdpr.org}, thus focusing on the language used rather than the users' identity.
Our work is a step towards creating effective \emph{hate speech} classification methods that require minimal input information to train a decision-making mechanism.

\section{Related work in hate speech classification}
\label{relWork}

The problem of \emph{hate speech} has drawn public attention, \cite{AlHassan2019}, with many research groups around the globe making efforts to address the issue systematically.
Existing solutions range from simple word-based approaches to complex artificial intelligence algorithms that employ Neural Network architectures.
The former applies NLP rules to input data, thus operating as supervised learning classifiers.

In general, NLP solutions exploit the lexical and syntactic features of sentences,
\cite{ChenZZX12}. 
In fact, for the problem of \emph{hate speech}, NLP solutions prove to be ineffective, as they fail to identify the subtle offensive content in user-generated text, mainly due to \emph{word ambiguity} and \emph{spelling variation} problems, \cite{schmidt-wiegand:2017}.
Moreover, the complexity of the natural language constructs, along with the fact that, quite often, there is a deliberate replacement of individual characters of offensive words with special ones to become undetected, are the two areas in which traditional unsupervised learning models fall short.

Using pre-trained word embeddings, such as \emph{GloVe}  by \cite{GloVe} or \emph{Word2Vec} by \cite{word2vec}, to encode input data is a fairly common technique used in NLP tasks, known to provide improved performance over random vector initialization.
In \emph{hate speech} tasks, this solution, despite the performance penalty due to the limited number of words that can be associated with pre-trained word vectors, is nevertheless adopted by the vast majority of \emph{hate speech} classifiers in the literature, \cite{meyer-gamback-2019-platform,Park2017,karatsalos2020,mahata-etal-2019}.

As far as the supervised learning classification solutions, in a work by \cite{DavidsonWMW17}, \emph{hate speech} in a short text is distinguished from the ordinary offensive language using Support-Vector Machine (SVM) and Logistic Regression (LR) classifiers.
Meanwhile, \cite{jha2017}, interested in detecting \emph{sexism} alone, have applied FastText, an algorithm by \cite{joulin2016bag}, along with SVM classifiers in their model.
Deep Neural Networks (DNN) is a type of supervised learning approach, able to solve problems end-to-end.
That is, they use labelled data to extract features from, and therefore they can handle text classification tasks quite well.
Convolutional Neural Networks (CNNs) and Recurrent Neural Networks (RNNs) are the two main architectures suitable for this task.
RNNs have short-term memory characteristics, and therefore they can handle dependencies in the input data sequences.
\cite{Gambck2017,Park2017} are some examples of works that use encoded input information in the form of characters and word n-grams to train CNN-based models.

Unlike CNN, Long Short-Term Memory (LSTM) is a type of RNN that captures long-term dependencies between words, and therefore it is made more suitable for text data.
The solutions by \cite{Badjatiya:2017,Pitsilis2018} are some of the works in the area of \emph{hate speech} detection, based exclusively on either LSTM or bi-directional LSTM (BiLSTM) architectures, \cite{Agwaral2019,Grosz20,qian-etal-2018}.
Gated Recurrent Unit (GRU) by \cite{cho-etal-2014} is another type of LSTM architecture that is becoming increasingly popular for a variety of classification tasks.
In NLP tasks, GRU-based networks perform similarly to LSTMs, but interestingly, they perform best only when applied to particular datasets with short text content, such as tweets, \cite{paetzold-etal-2019,mahata-etal-2019}.

To improve detection performance, other researchers have introduced hybrid approaches that combine CNN and RNN architectures of various types that are also worth mentioning,
\cite{siddiqua-etal-2019,meyer-gamback-2019-platform}.
In addition, some solutions incorporate Attention Mechanisms into LSTM-based designs, \cite{nikhil-etal-2018,Grosz20,Agwaral2019,karatsalos2020}.
Attention overcomes the limitations of LSTM, forcing any input sequences to be encoded into internal vectors of constant length, thus making this mechanism effective for very long sentences.
Finally, in another interesting work by \cite{Kapil:2020} that also falls into this category, knowledge from different trained models is used to enhance the classification performance of a multi-task LSTM-CNN model.

Bi-directional Transformers (BERT) encoder representations by \cite{Devlin2019BERTPO}, is a new method that integrates pre-training language representations into the classification mechanism in the form of sentence embeddings.
BERT embeddings were generated with bi-directional training of Neural Network, with the result that language representations are dynamically informed by the words around them.
Although successful in many text classification tasks, including \emph{hate speech} detection, 
\cite{bodapati-etal-2019-neural,swamy-etal-2019-studying}, BERT-enabled solutions also fall into the same category as those dependent on pre-trained word embeddings.

Numerous papers in the literature suggest solutions to the problem of \emph{hate speech} that involve more than one stage in the classification process.
From our standpoint, we consider a classification process as \emph{multi-stage} if there exist intermediate classification stages before the final one, which determines the class of a sample.
The models by
\cite{Park2017,siddiqua-etal-2019,risch2018}
fall into this category, employing two and three CNN and RNN classification components respectively, in the first stage.
In most examined cases of models of this category, the class of an input sample is determined by a single vector, resulting from the concatenation of all first stage output vectors.
In another, more sophisticated solution by \cite{meyer-gamback-2019-platform}, the output of the first stage is merged through a dense layer to form the classification result of the second stage.

In addition, there is a specific category of multi-stage classifiers, which incorporate \emph{ensemble learning} into the classification process.
Based on the principle of combining multiple low-performance classifiers to produce a high-performance model, ensemble learning has proven effective in most machine learning tasks.
In this learning scheme, the first stage employs several classifiers based on various neural network architectures such as CNN, RNN, GRU, etc., or classic ones such as XGBoost, ADAboost, LR.
The output of these is then combined in the next stage to determine the class of the text input.
This is performed either through some form of voting, \cite{rajendran-etal-2019-ubc,madisetty2018,liu-etal-2020-scmhl5,seganti-etal-2019-nlpr} or by averaging the classification output of the first stage, \cite{zhang2019identifying}.

The problem of detecting aggressive language that uses out-of-vocabulary words has also been approached by several researchers,  \cite{madisetty2018,raiyani-etal-2018-fully,liu-etal-2020-scmhl5}, distinguishing the \emph{Overtly Aggressive} hateful textual content from the \emph{Covertly Aggressive}.
Nevertheless, no results exist so far to demonstrate any performance gain over the classic approaches, nor are there many differences between the strategies used to detect direct and indirect aggression.

As for the language the hatred content is written in, most of the solutions in the literature refer to methods whose effectiveness has been demonstrated only for English text content.
Moreover, a small number of works that deal with the \emph{hate speech} problem in other languages, such as Portuguese, German, Arabic, Hindi, Italian, use algorithms adapted to the specifics of each one.
Nevertheless, to the best of our knowledge, no method exists today to provide language-agnostic features, mainly due to the limitations imposed by the employment of pre-trained word embeddings.

A summary of the existing solutions to \emph{hate speech} classification, along with their characteristics, is presented in table \ref{cartography}.
Column 2 refers to the central components of each presented solution.
In column 3 is shown the variety of classes of \emph{hate speech} that can be detected, while column 4 shows the number of classification stages involved in each algorithm.
Column 5 lists the datasets used in model evaluation, with \emph{twitter} being the most popular.
In column 6 is shown the type of word embeddings used in each approach.
As shown in the table, most of the approaches listed are dependent on pre-trained word embeddings.
Finally, the last column provides additional information on user characteristics employed in model training.

It is also interesting to note the inherent difficulty of the \emph{hate speech} challenge itself, stated by the fact that no solution, thus far, has been able to obtain an \emph{F-score} above 0.93 overall, nor above 0.70 for the \emph{racism} class alone, in multi-class datasets, such as by \cite{Waseem2016b}.
Despite the plethora of solutions to the \emph{hate speech} problem published so far, we believe that further performance improvements are still feasible through proper fine-tuning and alterations to the existing algorithms.

\begin{table*}

\begin{adjustwidth}{-0.5in}{-0.5in}
\centering
\caption{Cartography of existing research in hate speech detection}
\label{cartography}
\scalebox{0.80}{ 
\begin{tabular}{|c|c|c|c|c|c|c|}
\hline
 & & & & & & \textbf{Additional} \\
  \textbf{Citation} & \textbf{Model} & \textbf{Classes to detect} & \textbf{Classif.} & \textbf{Dataset} & \textbf{Embed.} & \textbf{user} \\                           
 & & & \textbf{stages} & \textbf{used} & \textbf{used} & \textbf{features} \\
\hline
\hline
\multirow{9}{*}{
}
 \mbox{
\citet{ChenZZX12}
}
&    NLP   &  continuous values      &  1 &  YouTube  & N/A & Conversation \\ 
 & Unsupervised & & & & & history \\
\hline

\mbox{
\citet{Waseem2016b}
}
 & Logistic Regression & 3:(Sexism,Racism,None) & 1 & twitter & N/A & Gender, \\ 
 &          &        &   &  & & Location \\

\hline

 \mbox{
\citet{paetzold-etal-2019}
}
 & RNN - GRU & 2:(Hate,Non-Hate) & 1 & twitter & CharToWord & N/A \\
                            &     & 3:(Group,Individual,Other)      &   & & & \\
\hline

 \mbox{
\citet{Kapil:2020}
}
 & CNN - LSTM & 2:(Offence,Non-offence) & 1 & twitter & Word2Vec & N/A \\
   &  GRU    & 3:(Hate,Offence,Neutral)  &   & & &\\
   &         & 3:(Racism,Sexism,Neutral)  &   & & &\\
   &         & 3:(Overtly,Covertly,None)  &   & & &\\

\hline

 \mbox{
\citet{DavidsonWMW17}
}
 & SVM - Bayes - Log.Reg. &  3:(Hate,Offensive,Neither)  &  1     & twitter & N/A & N/A \\ 
                & Decision Trees - Ran.For. &   & & & &\\
\hline

 \mbox{
\citet{jha2017}
}
 & SVM - FastText & 3:(Benevolant,Hostile,Other)   &   1   & twitter & N/A & N/A \\
 &          &           &      &   &  &  \\
\hline

 \mbox{
\citet{Badjatiya:2017}
}
&  CNN - LSTM  & 3:(Sexism,Racism,Neither)  &  1   & twitter & GloVe & N/A \\
                               & FastText &  &  & & & \\
\hline

\mbox{
\citet{Park2017}
}
 & CNN    & 3:(Sexism,Racism,Neither)   &  1 - 2  & twitter & Word2Vec & N/A \\ 
 &          &           &      &   &  &  \\
                         
\hline
 \mbox{
\citet{Gambck2017}
}
& CNN & 4: & 1     & twitter & Word2Vec & N/A \\ 
                         &        & (Sexism,Racism,Both,Neither)   &      & & & \\
\hline

\citet{Pitsilis2018}
& LSTM    & 3:(Sexism,Racism,Neither) & 1     &  twitter &  N/A & Tweeting \\ 
                         &        &    &      &  & & history \\
\hline

\citet{meyer-gamback-2019-platform}
&  CNN - RNN   & 3:(Sexism,Rasism,Neither)  &  2    & twitter & GloVe &  N/A \\ 
                         &   LSTM     &   & & & Word2Vec & \\

\hline

\citet{swamy-etal-2019-studying}
&  SVM - LSTM  & 3:(Sexism,Rasism,Neither)  &  1    & twitter & GloVe &  N/A \\ 
&   BERT     & 3:(Hateful,Offensive,Neither)  & &  & BERT & \\
&            & 2:(Hate,Non-Hate)  & &  &  & \\

\hline

\citet{karatsalos2020}
& RNN - Attention & 4: &   1  &  twitter & GloVe & N/A \\ 
  &   & (Sexual,Physical,Indirect,None)     &      &  &  & \\
\hline

\citet{Agwaral2019}
& Bi-LSTM    & 2:(Attack,Non-Attack) classes &  1    & wikipedia & GloVe & N/A \\ 
            &   Attention     &  3:(Sexism,Racism,Neutral) &      & twitter & SSWE & \\
\hline

\citet{Grosz20}
&  Bi-LSTM   & 2:(Sexist,Non-Sexist) &   1   & twitter & GloVe &  N/A \\ 
&          &           &      &   &  &  \\
\hline

\citet{qian-etal-2018}
&  Bi-LSTM  & 2:(Hate,No-Hate) &   1   & twitter & N/A &  Tweeting \\ 
               &          &           &      &   &  & history \\
\hline

\citet{siddiqua-etal-2019}
&   CNN - LSTM  &   2:(Hate,Non-Hate) &  2  & twitter & FastText & N/A \\ 
&          &           &      &   &  &  \\
\hline

\citet{nikhil-etal-2018}
& LSTM - Attention & 3:(Overtly,Covertly,None) &   1   & Facebook & N/A & N/A \\
&          &           &      &   &  &  \\
\hline

\citet{bodapati-etal-2019-neural}
&  CNN - BERT   & 2:(Toxic,Non-Toxic) &   2   & wikipedia & Word2vec & N/A \\ 
&  FastText & 3:(Sexism,Racism.Neither)  &      & twitter & BERT & \\
                         
\hline


\citet{rajendran-etal-2019-ubc}
& Log.Reg. - SVM - Ran.For. & 2:(Targeted,Untargeted) &      & twitter & N/A & N/A \\ 
 & ADABoost - XGBoost &  3:(Individual,Group,Others) &      &  &  & \\
\hline

\citet{madisetty2018}
& CNN - LSTM    & 3:(Covertly,Overtly,None) &   1   & Facebook & GloVe & N/A \\ 
                         &  Bi-LSTM      &   &      &  &  & \\
\hline

\citet{liu-etal-2020-scmhl5}
&  BERT   & 2:(Gendered,Non-Gendered) &   1   & Facebook & BERT & N/A  \\ 
  &     &  3:(Covertly,Overtly,None)         &      &  &  & \\ 
\hline

\citet{seganti-etal-2019-nlpr}
& LSTM - SVM    & 2:(Hate,Non-Hate) &   1   & twitter & FastText & N/A \\ 
 &  Random Forest      &   &      & wikipedia &  & \\

\hline

\citet{mahata-etal-2019}
& bi-LSTM - CNN        &  2:(Offensive, Non-offensive) &  2 & twitter & GloVe & N/A \\ 
           &   bi-GRU &   &    &         &     & \\
\hline

\citet{raiyani-etal-2018-fully}
&  Log.Reg. - LSTM   & 3 classes aggression &   1   & Facebook & GloVe & N/A \\ 
     & CNN - FastText & (Overtly,Covertly,None) &     & twitter &  & \\
\hline

\citet{mishra-etal-2018}
&  GRU - LSTM  & 3:(Sex,Rac,Neither) &   1   & twitter & GloVe & N/A  \\ 
&   CNN   & 3:(Toxic,Attack,None)  &      & wikipedia & Char-based & \\
\hline

\citet{fehn-unsvag-gamback-2018}
&  Logistic Regression  & 2:(Hate, Non-Hate) &   1   & twitter & N/A & Behavioral \\ 
                       &        &     &      &  &  & data \\

\hline
\end{tabular}
}
  \end{adjustwidth}
\end{table*}

\section{Proposed Framework}
\label{model}
This section describes the system architecture for the proposed two-stage \emph{hate speech} detection approach.
As mentioned above, to take advantage of the multi-label data, in our approach, we apply the concept of \emph{multi-label classification} in the first stage.

The concept of multi-label classification is not new, but it is becoming increasingly popular in modern applications, \cite{Tsoumakas09}.
\emph{Problem transformation} is a sub-category of this classification concept.
According to this, the initial problem of classifying samples on multi-label data is transformed into a series of classification problems on single-label data.
This approach is beneficial for data with overlapping classes in the feature space, (i.e. labels not mutually exclusive).

To apply this concept of multi-label classification to our particular case, we first convert our multi-label data into multiple single-label datasets, with each one then used to train an individual OvR classifier.
By this step, a series of intermediate classification output is produced.
The intermediate classification result produced by these classifiers is then incorporated into the second stage, where the final category of a given sample is determined.
Also very important for our task is to capture any indirectly expressed toxicity.

\subsection{Capturing Indirect Toxicity}

According to other scientists in the field, \cite{fernandez2018}, it is very common for aggressive language to be expressed indirectly, e.g. in the form of \emph{sarcasm}.
That usually happens when using descriptive language containing either out-of-vocabulary words or just common phrases.
Capturing such text expressions requires the modelling of additional linguistic information, such as likely dependencies between strongly connected words in the sentence, based on the principle that: the next word appears according to the word before it.
For the detection of \emph{racism}, however, such modelling seems very promising because it is usual in verbal communication to convey such a form of \emph{hate speech} without necessarily using toxic words.
Using word n-grams is one way to capture the above requirement.
As also observed by \cite{Nobata2016}, using any n-gram word-based features, can improve performance in the task of detecting \emph{hate speech} in English, rather than using only word-based features.
Inspired by this, in our design, along with features based on word uni-grams, we also incorporate word bi-grams and tri-grams, thus capturing the sequence of adjacent elements as additional features.

\subsection{First stage classification}

Let $D$ be the set of samples and $L=\{l_S,l_R,l_N\}$ the labels corresponding to these samples.
In a single, multi-label classification scheme that works on three-label data, for the output class of a sample $x \in D$, the following holds:
$H : x \rightarrow \{ p_{_S}, p_{_R}, p_{_N} \}$, with $p_{(i)}$ the predicted probability for sample $x$ to receive label $l_{(i)}: i \in \{S,R,N\}$.

In our problem transformation scheme, the initial set $D$ is converted to $|L|$ sets, with each one corresponding to the labels ”Racism”, ”Sexism”, and ”Neutral”, respectively.
Each new set includes all the samples of the original one $(|D_S|=|D_R|=|D_N|=|D|)$, but with their labels converted to the relevant binary value.
As a result, the training scheme then involves $|L|$ binary classifiers instead of just one, so that the output for the class label $l \in \{l_S, l_R, l_N\}$ of a given sample $x$ is as follows:
$H_S : x \rightarrow (p_{_S},p_{\neg _S} )$,
$H_R : x \rightarrow (p_{_R},p_{\neg _R} )$, 
$H_N : x \rightarrow \{p_{_N},p_{\neg _N} \}$
with $\neg p_{(i)}=(1-p_{(i)}) : i \in \{S,R,N\}$.

\begin{figure*}
\centering
\hspace*{-0.1in}
\includegraphics[width=.80\textwidth]{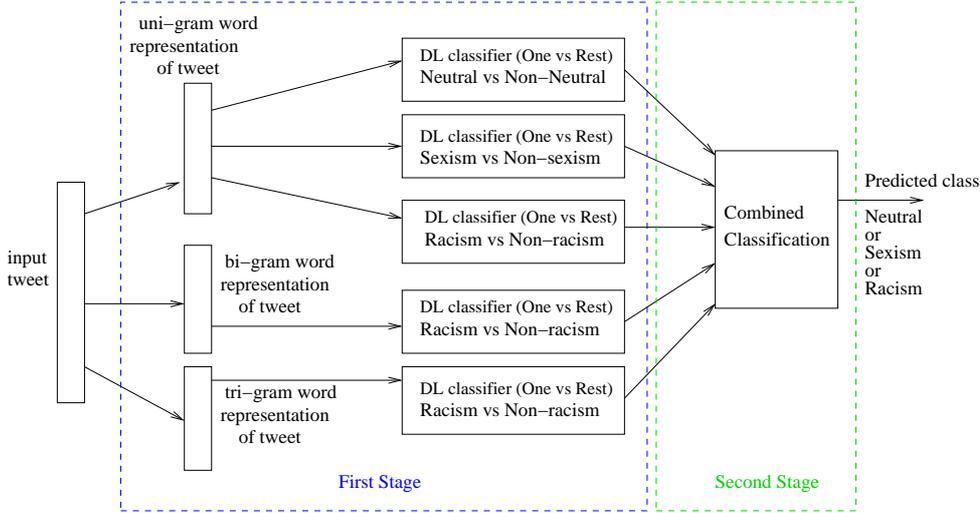}
\caption{High level view of the proposed architecture}
\label{fig:fstStageB}
\end{figure*}

To serve the additional requirement to detect, through the classification scheme, possible indirect toxicity expressed in \emph{racism}, we incorporate two other binary classifiers into our design.
To extract additional features from word bi-grams and tri-grams requires to have two other datasets $D_{R_2}$, and $D_{R_3}$, respectively, derived out of the original set D, with sizes equal to: $|D_{R_2}| = |D_{R_3}| = |D|$.
As with standard classifiers, for the additional ones, the following holds for the class output of a given sample $x$: $H_{R_2} : x \rightarrow (p_{_R}, p_{\neg _R})$ and $H_{R_3}: x \rightarrow (p_{_R}, p_{\neg _R})$.
The high-level view of our proposed architecture is illustrated in fig.\ref{fig:fstStageB}.\\

\begin{figure*}
\centering
\hspace*{-0.1in}
\includegraphics[width=.90\textwidth]{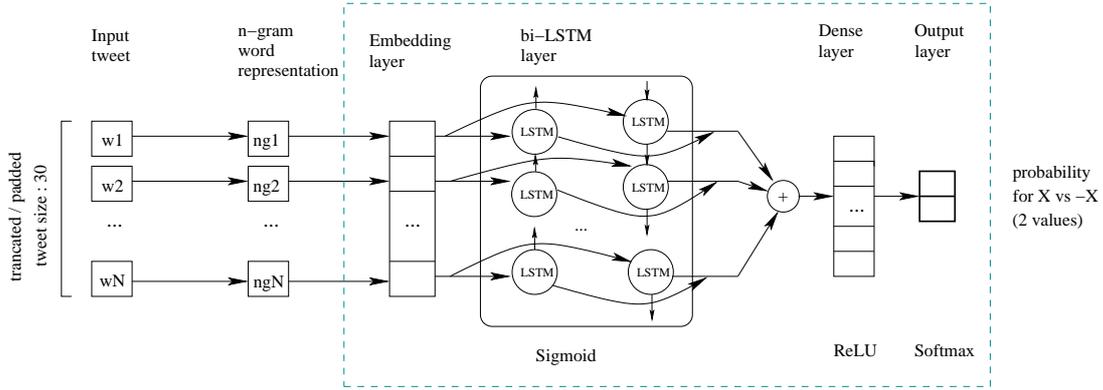}
\caption{The typical One-vs-Rest (OvR) Deep Learning classifier model used in the first stage}
\label{fig:fstStageA}
\end{figure*}

In our design, we chose to apply a single Deep Learning model to all of the first stage binary OvR classifiers described above.
Having to deal with sequential data such as text, we chose to employ an LSTM-based model.
As with RNNs, \emph{Sequential Memory} is a key feature of LSTMs, making this type of neural network suitable for text sorting.
This is because, in sequential memory, a signal that passes to the next neuron depends, to a certain degree, on the content of the previous neurons in the network, \cite{lipton2015critical}.
That makes it possible to capture the sequence of words in a sentence.
Unlike RNNs that suffer from the \emph{vanishing gradient} problem, LSTMs have longer short-term memory, and therefore improved ability to retain longer sequences.
In addition, to achieve a substantial performance improvement, we have selected an enhanced type of LSTM, called bidirectional LSTM (bi-LSTM), which contains both forward and backward layers.
That is, one layer, which is a reversed copy of the other, allows the neural network to look into words occurring both before and after the current one.
For our design, we chose to implement a uniform Deep Learning architecture, shown in fig.\ref{fig:fstStageA} and apply it to all OvR first stage classifiers.
As such, their input vectors can consist of either word uni-grams,bi-grams or tri-grams.
This architecture comprises four layers, described as follows:\\

\begin{itemize}
  \item \emph{Input (a.k.a Embedding) layer}. 
  The \emph{input dimension} determines the vocabulary size, with its value set to 25k, 120k and 180k for the case of uni-grams, bi-grams and tri-grams input, respectively.
  The selected values approximate the number of unique n-grams in the input datasets.
  Also, the size of the \emph{input sequence} is set to 30, indicating the maximum number of words the system can handle for each sample of input text.
  
  \item \emph{The Hidden bi-LSTM layer}.
  The bi-directional LSTM is the main layer for processing text sequences of any type of word n-grams, and it is fully connected to both the \emph{Input} and the subsequent layer.
  Based on preliminary experiments, the output space was set to 30 so that it is equal to the number of features on the \emph{Embedding layer}.
  \emph{Sigmoid} is the chosen activation function.

  \item \emph{Dense layer}.
  This layer was added to receive the output of the bi-LSTM layer and improve the learning.
  The size of this layer was set to 30 so that it is equal to the input text size on the \emph{Embedding} layer.
  \emph{ReLU} (Rectified Linear Unit) is the activation function used.
  
  \item \emph{The Output layer}.
  This layer consists of only 2 neurons to provide output in the form of probabilities for the $l$ and $\neg l$ class.
  Consequently, \emph{softmax} was chosen for this layer, as the most suitable activation function for this type of output.
\end{itemize}

For the training process, \emph{binary-cross-entropy} is the loss function chosen as the most suitable for two-class output.

\subsection{Second Stage Classification}
\label{SecondStage}
The general idea is to use the predictions generated over the first stage, as features for the training of the second stage classifier.
In fact, for a given sample $x$, the output from each first stage OvR classifier has the form of two-tuple $(p_{(i)}, \neg p_{(i)})$, for which holds: $\neg p_{(i)}=(1 - p_{(i)})$,
with $i \in \{1,..,c\}$ denote as the classifier, and $c \in\{3,5\}$.
All two-tuples created are then combined into a single vector with concatenation.
Next, the label initially given to sample $x$ is attached to the concatenated output, thus creating the input features for the second stage classifier.
The resulting vector has either a 9-tuple or 13-tuple form $T:(p_1,\neg p_1,p_2,\neg p_2,...,p_i,\neg p_i,B_S,B_R,B_N)$, depending on the actual number of OvR classifiers selected in the scheme.
The correspondence between the initial label of a sample and the values $B_S,B_R$ and  $B_N$ is given below, with  $B_{(d)}: d \in \{l_S,l_R,l_N\}$:
\begin{equation}
    (B_S,B_R,B_N) = 
\begin{cases}
    (1,0,0) : Sexism\\
    (0,1,0) : Racism \\
    (0,0,1) : Neutral \\
\end{cases}
\end{equation}

The final class $H_{class}$ determined for a new sample $x$ in the second stage is derived by a function $\mathlarger{f}()$, as described in equation \ref{E0}.
$Y$ denote as the set of the classification output values $y$, generated in the first stage.

\begin{equation}
\label{E0}
H_{class}(x) = \mathlarger{f} ( \underset{y \in Y}{\mathlarger{\cup}} \{y\} : H_i(x) = y )
\end{equation}
In our comparative study, we tested various alternative algorithms suitable for the task of the second stage.
These include several conventional \emph{boosting} \& \emph{bagging}, such as: \emph{Logistic Regression}, \emph{Random Forest}, \emph{Adaptive Boosting}, \emph{Gradient Boosting} and \emph{Extreme Gradient Boosting}.
We also proposed and tested a Deep Learning architecture that combines the output of all first stage classifiers.
Finally, we introduced a simple ensemble scheme, which employs a set of fixed criteria to implement the function $\mathlarger{f}()$, as described in more detail below.

\subsubsection{Logistic Regression (LR)}

We chose to include this algorithm in our experimentation because it is a simple, fast, and efficient classification method.
LR is a statistical model which uses, in its primary form, a logistic function to model a binary dependent variable.
Variants of this model are also suitable for the classification of non-binary dependent variables.

In our scheme, the output generated by the first stage OvR classifiers becomes the input variables in the LR.
In our particular case, we limited our experimentation to the following loss functions only due to their suitability to produce multi-class output:
i)  \emph{Stochastic Average Gradient}.
ii) \emph{Limited-memory BFGS}. 
iii) \emph{Newton Cost Function}.
To optimize the result, we tried all the above loss functions in every fold, from which we finally selected the best one.

\subsubsection{Random Forest (RF)} 
This one is a variant of a bagging algorithm by \cite{randomForest}.
As its name implies, in Random Forest, multiple decision trees are formed and run in parallel on a different sample of data, making up a forest of classifiers that work as an ensemble.
Each sample is a sub-sample from the original data, provided as input to the second stage of classification.
In this way, a low correlation is achieved between the individual decision trees, automatically correcting the output from the individual trees and thus minimizing the error.
The class prediction of each tree is taken into consideration in a voting scheme to decide the finally determined class of an input sample.

We experimented with various numbers of trees in the forest, as well as with tree depths, ranging from 1 to 100 and 10 to 30 respectively.

\subsubsection{Adaptive Boosting (AdaBoost)}

In contrast to bagging algorithms and their limitations imposed by the parallel training, in boosting algorithms instead, several low-performing weak learners are employed and trained sequentially.
In AdaBoost, an algorithm by \cite{FreundS96}, the classification decision evolves via a voting scheme among learners. 
The errors of the data points are re-weighted through iterations,  giving greater weight to those not well classified.

In its general form, the output class, for two-class classification problems, is provided by the following equation:

$F(x)=sign(\mathlarger{\sum\limits_{m=1}^{M}} \theta_m f_m(x))$ \\

where $f_m$ denote as the learner $m$ out of the $M$ used, and $\theta_m$ the corresponding weight to that learner.

While the standard algorithm refers to binary classification, for our three-class problem, we chose a variant that incorporates an ensemble of binary decision trees (BDT) and is, therefore, suitable for multi-class output.
Simply put, a multi-class problem breaks down into several binary classification problems solved by standard AdaBoost classifiers,
\cite{Fleyeh2013}.
To reduce overfitting, we experimented with a range of \emph{learning rates}, from 0.01 to 0.14, and various binary trees from 5 to 50 to optimize the algorithm.

\subsubsection{Gradient Boosting (GBoost)}

Like AdaBoosting, Gradient Boosting  by \cite{Friedman2001}, combines weak learners, but without limiting the depth of the Decision trees to just one.

The objective of the first iteration is to fit the model to the complete data.
Then, in the subsequent iterations, a new weak learner is added to the model to increase the performance and thus build a strong learner.
The new weak learners concentrate on those areas where the performance was still poor in the current step until the model fits the data exactly.

In our scheme, we chose the \emph{deviance} loss function for its suitability for multi-class classification.
We also experimented with a range of \emph{learning rates}, from 0.10 to 0.20.
In addition, we tested the algorithm for a range of binary trees from 20 to 100, while limiting our experimentation to decision trees to depths of 2 and 4 only, from which we selected the best combination.

\subsubsection{Extreme Gradient Boosting (XGBoost)}

XGboost algorithm by \cite{Chen2016}, is a variant of Gradient Boosting, but improved for speed and performance, which uses the second order derivatives of the loss function to minimize the error.
Also, the binary decision trees in XGBoost can have a varying number of terminal nodes.
The advanced regularization of this type which improves generalization is another advantage over the standard Gradient Boosting.
We experimented with a range of binary trees from 20 to 30, with a maximum depth of 2, and \emph{learning rates} ranging from 0.1 to 0.22, from which we chose the combination with the best performance.

\subsubsection{Deep Learning (DL) with Stacked Ensemble}
\label{DLStackEnsemble}
 
Deep neural networks, among other uses, have also been found suitable for classifying patterns of sequences of numbers, \cite{Gao2018}.
In this regard, we are considering using LSTM-based Deep Learning models for the second stage classifier training task.

\begin{figure}[!ht]     
   \centering
   \hspace*{-0.1in}
   \includegraphics[width=.90\textwidth]{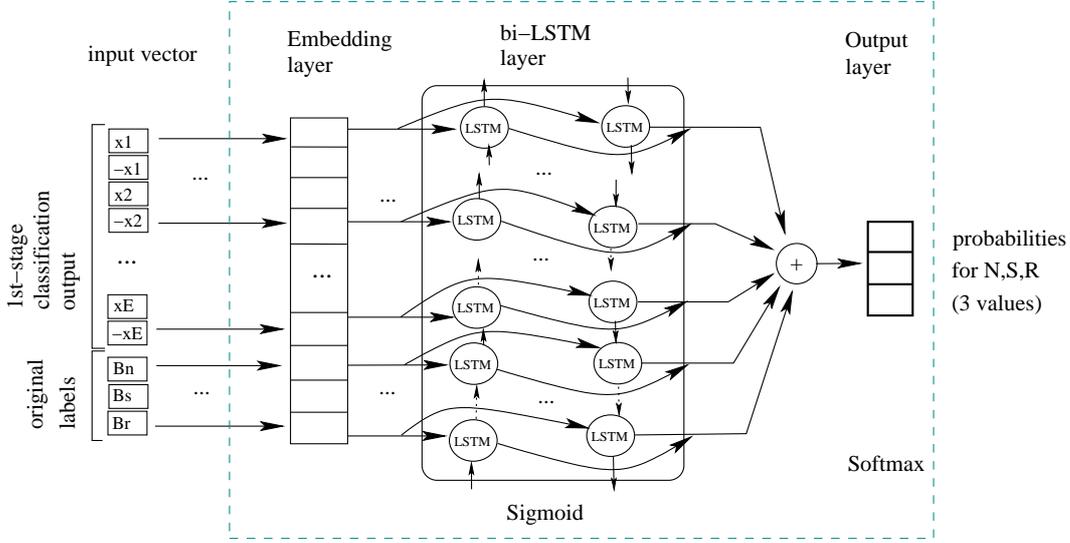}
   \caption{The DL-based alternative of the algorithm used in the second stage. $E$ refers to the number of 1st stage classifiers used.}
   \label{fig:2ndStageLSTM}
\end{figure}

The LSTM-based architecture employed in our design is illustrated in fig.\ref{fig:2ndStageLSTM} and comprises three layers, described as follows:

\begin{itemize}
    \item \emph{Input (a.k.a Embedding) layer}. 
    The input dimension of that layer is set equal to the vector space dimension of the first stage output i.e.: $c \times 2 \times 10^m$, where $c$ denotes the number of binary OvR classifiers incorporated into the second stage, and $m$ is the number of numerical digits used for expressing each output probability for the classes $l_{(i)}$ and $\neg l_{(i)}$.
    In other words, $m$ indicates the number of different states which the output of each first stage classification can take, and which the second stage classifiers trained to identify.

    \item The \emph{Hidden bi-LSTM layer}.
    The bi-LSTM is the main layer for processing the concatenated output sequences generated in the first stage.
    Based on preliminary experiments we set the output space dimension to 100. \emph{Sigmoid} is the activation function chosen for this layer.
    
    \item The \emph{Output layer}.
    This layer consists of 3 neurons and provides the output probabilities for the three classes of \emph{neutral}, \emph{sexism} and \emph{racism}.
    The \emph{softmax} activation function is chosen for this layer to generate multi-class output.
\end{itemize}

\begin{figure}[!ht]    
   \centering
   \hspace*{-0.1in}
   \includegraphics[width=.70\textwidth]{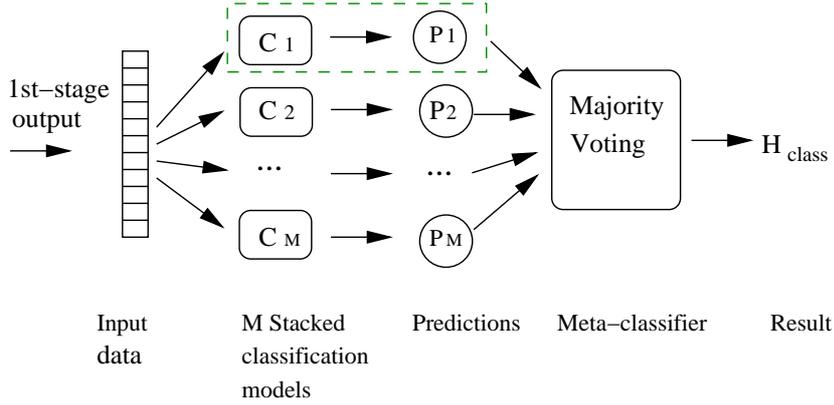}
   \caption{The stacked ensemble scheme. The elements surrounded by dotted line correspond to DL-based classifier shown in fig.\ref{fig:2ndStageLSTM}}
   \label{fig:stackedLSTM}
\end{figure}      

\emph{Categorical cross-entropy} is the loss function chosen for the training process, as it is suitable for generating multi-class output.
The decision to choose \emph{bi-LSTM} for the hidden layer was made based on performance criteria, with \emph{bi-LSTM} being the scheme of neural network that produces superior classification accuracy among others that we experimented with, such as CNN and GRU.

Moreover, many submodel instances of the above DL-based architecture in a stacked ensemble layout make up the second stage classification scheme shown in fig \ref{fig:stackedLSTM}.
In Generalization Stacked Ensembles, \cite{Ting_1999}, for a given DL model, several identical sub-models are created and trained on exactly the same data, within the same set of parameters.
Although requiring longer training times, Generalization Stacked Ensembles are more beneficial in terms of classification performance, than a single-trained instance.

For the classification of new samples, the predictions made by the trained sub-model instances of the Stacked Ensemble scheme are finally combined into a single decision.
In our particular case, the predicted class for a given sample is determined via a \emph{Majority Voting} scheme, which combines the classification output from all first stage classifiers, as shown in eqn.\ref{form:stacked}, and it works as follows:

\begin{equation}
\label{form:stacked}
H_{class}(x) = argmax( \mathlarger{\{}P: \underset{j \in \{l_R,l_S,l_N\}}{P_j}=\sum_{i=1}^{M}k_{i,j}\mathlarger{\}} )
\end{equation}

Given a model with $M$ stacked instances, then for an input sample $x$, the predicted frequencies for the \emph{neutral}, \emph{sexism} and \emph{racism} class is calculated by summing the adjacent frequencies across all $M$ instances.
Next, the $H_{class}(x)$ label finally determined for sample $x$ is given to that class $j \in \{l_S,l_R,l_N\}$ having the highest summed frequency.
$k_{i,j} \in \{0,1\}$ denote as the predicted probability for class $j$, computed by the ensemble instance $i$, while $P$ is the cumulated prediction score over all $P_j$ values for sample $x$.

\subsubsection{Fixed Criteria Ensemble scheme}
\label{intuitive}

We propose an alternative scheme for application in the second stage for determining the class of textual postings.
In this scheme, we employ a simple and intuitive mechanism, based on fixed criteria, to aggregate the output of all first stage OvR classifiers into a single score.
In total, we propose two forms of the scheme, for comparison.
We apply this dual-form testing as a means to investigate the effectiveness of the proposed scheme on capturing the indirect toxicity expressed over \emph{racism}.
While the first one considers only the word uni-grams criteria, in the second one, the classification scheme incorporates the word bi-grams and tri-grams along with the uni-grams.

The first form is described by a function $\mathlarger{h}(): x \rightarrow \{ l_S, l_R, l_N \}$, we introduce in eqn.\ref{E1}, which determines the class label of an input sample $x$.
$p_N, p_S$ and $p_R$ denote as the positive class probability values of the prediction scores obtained for \emph{neutrality}, \emph{sexism} and \emph{racism} by the relevant first stage classifier, respectively. 
The intuition behind function $\mathlarger{h}()$ is to assign to sample $x$ the class label for which the adjacent first stage classifier has obtained the highest predicted score for that sample.

\begin{equation}
\label{E1}
H_{class}(x)=\mathlarger{h}(max\{p_{_N}(x),p_{_S}(x),p_{_R}(x)\})
\end{equation}

\begin{equation}
\label{E2}
H_{class}(x)=\mathlarger{h}(max\{p_{_N}(x),p_{_S}(x),avg\{p_{_R}(x), p_{R_2}(x), p_{R_3}(x)\}\})
\end{equation}

The second form of the proposed scheme is shown in eqn.\ref{E2}. 
The most obvious difference with the previous one is the replaced $p_R$ class probability by the average of the three probabilities values of the \emph{racism} class: ($p_{R_2}(x)$,$p_{R_3}(x)$,$p_R(x)$).
The purpose is the inclusion of word bi-grams and tri-grams, along with the uni-grams in the scheme.
The reasoning behind selecting the average value was to consider, equally in the class decision, all features related to \emph{racism}.

\section{Evaluation}
\label{eval}

To demonstrate the effectiveness of the proposed classification algorithm, we implemented the scheme shown in fig.\ref{fig:fstStageB} and evaluated it over a legacy dataset suitable for this task.
Furthermore, to test how well the model generalizes, we conducted experiments of the trained models over another public dataset, which we used as the test set.
For both datasets, we first applied the necessary pre-processing, as further explained in this section below.

\subsection{Dataset}
\label{dataset}

To evaluate the model, we used an existing dataset with approximately 16,000 tweets by \cite{Waseem2016b}.
This data set is an established and accepted standard for this evaluation task and it is constantly cited as a valuable resource.
That makes it suitable for producing comparable evaluation results against other state-of-the-art solutions.
Moreover, other researchers, such as  \cite{mishra-etal-2018-neural}, who analyzed the textual content of this dataset, report the existence of a large number of tokens that are not included in the English dictionary, which, however, appear mainly in \emph{racist} and \emph{sexist} tweets.
More specifically, 5.2 thousand of the 16 thousand unique tokens found in the dataset do not exist in the dictionary.
This feature makes the data set ideal for investigating any indirectly expressed \emph{hate speech}.

Using the public Twitter API, we were able to retrieve 15998 tweets, of which 1943 labeled as \emph{racism}, 3166 labeled as \emph{sexism}, and 10889 labeled as \emph{neutral} (i.e., tweets that do not contain either \emph{sexism} or \emph{racism}).
The discrepancy observed regarding the quantities between the data we retrieved and the original set is because a number of tweets have been removed from the Twitter service.

\subsection{Pre-processing}

Before training the first stage classifier, it is necessary to apply the proper pre-processing to the labeled input tweets.
That includes Tokenization, Encoding, and Data Augmentation, which we describe in the following sections.

\subsubsection{Tokenization}

Tokenization is the process by which the text corpus is split into word elements, taking into account white spaces as well as various punctuation symbols used in the language. 
For achieving this, we used the Moses\footnote{http://www.statmt.org/moses/} package for automatic translation.
It is worth mentioning that emojis, special characters and other
symbols, such as "@" used in email addresses, were retained in the dataset as we intended.
The objective was to train the model with the most realistic, high-quality data, as these symbols are often used on social media to obscure \emph{hate speech}.

\subsubsection{Encoding}
\label{encoding}
To fulfil our initial goal of maintaining independence from any linguistic rules and vocabularies, we choose to model the input tweets in the form of vectors using Word Frequency Vectorization.
We apply this technique to associate each word in the corpus with a unique identifier.
More particularly, in this type of vectorization, each word is indexed based on the frequency of appearance in the corpus.
The index value given to each word in a tweet is used as a vector element to describe that tweet.
Despite the potential loss of generalization, this modelling provides the flexibility to support, in theory, messages posted in any language.

To encode the attributes of the word bi-grams and tri-grams for each tweet, we apply a similar scheme of frequency vectorization.
First, we generate two more corpora, one for the word bi-grams and another for the tri-grams.
With this, a new unique identifier is assigned, respectively, to each two-word and three-word sequence appearing in the original corpus. 
In this way, each input tweet can be described as a vector in the bi-gram and tri-gram spaces.
Speaking in numbers, in the evaluation dataset, the corpora of uni-grams, bi-grams and tri-grams finally comprise 25k, 120k, and 180k unique words, respectively.
Unlike other models that use pre-trained word embeddings for text encoding, any toxic expressions such as "son of a b****" that contain out-of-vocabulary words, can still be classified as such by our approach.
The choice of vectorizing the sentences into n-grams, advances our proposed model so that it can tolerate similar cases of obfuscated \emph{hate speech}.
More important, this feature is provided in a free of pre-trained word embeddings scheme.
Finally, we choose to limit the maximum size of encoded tweets that will be taken into account in the model training to only 30 words.
As a result, any tweets longer than this size were truncated, while any shorter ones were padded with zeros.
The same rule applies to all tweet embeddings associated with word uni-grams, bi-grams and tri-grams, limiting their size to 30 elements.
This value is marked as 'input dimension' at the embedding layer in all first stage OvR classifier models.

\subsubsection{Data Augmentation}

Augmenting the training data is often necessary when the class distributions are highly imbalanced.
Otherwise, using an unbalanced data set usually has an undesirable impact on the calculated \emph{F-score}, with its value being strongly biased towards the majority class.
That produces many false negatives, which is undesirable in algorithm evaluation.
To give a figure of the level of imbalance occurring in the original dataset, we mention that the ”neutral” class alone accounts for 68\% of all samples.
For our proposed scheme, we look into the employment of data augmentation for both stages of the classification task.

As for the first stage of our scheme, possible options include the implementation of a simple data replication scheme or the employment of more advanced techniques, including a machine translation service, \cite{karatsalos2020,risch2018}.
However, the latter option requires the involvement of third-party language services in the scheme, with high dependency on linguistic rules and vocabularies, and is therefore not recommended for our problem.
For this reason, we chose the solution of data replication, which is much simpler.
With replication, to achieve a proper balance between the classes, the minority classes are over-sampled, using the original data.
The augmented data set is then appended to the original set, a practice that is quite common in \emph{hate speech} literature,
\cite{seganti-etal-2019-nlpr,Agwaral2019}. 
It is important to note that, in data augmentation, over-sampling should only be applied to the training set, to avoid over-fitting in the resulting data.
For the first stage, we over-sampled the classes of "sexism" and "racism", twice and thrice respectively.
As a result, these classes increased in size to 6,332 and 5,829 samples, respectively.
After incorporating the augmented toxic classes into the original 10,889 tweets of the "neutral" category, the training set finally expands to 23,050 samples.

In the second stage, the necessity for data augmentation stems from the fact that the minimal data produced in the first stage on each fold is rather inadequate for training properly the second stage classifiers.
%
%
In fact, as mentioned earlier in section 4.3, using a limited dataset to train the second stage classifier would almost certainly result in high instability and overfitting.
Using a simple oversampling scheme identical to the first stage, on the other hand, would most likely be insufficient to address the overfitting problem. 
For that reason, large-scale augmentation appears to be a viable option for our setup to achieve notable performance improvement.

In our augmentation scheme, new artificial samples are created for each class within the same dimensional space while maintaining a high similarity to the existing ones. 
Algorithm \ref{oversampling} explains the technique in detail. 
The set of samples belonging to the specific class (i.e., "sexism," "racism," or "neutral") that we wish to augment is denoted by $<$ \emph{original dataset} $>$, and $N$ denote as the new set of data generated by the algorithm. 
The Maximum Divergence Value (mdv) is a parameter used in data generation to determine the preferred level of similarity between class probability values within the same class.
\begin{algorithm}
\caption{Generation of new samples of a particular class}
\label{oversampling}
\begin{algorithmic}[1]
\Procedure{Gen\_data}{$n\_samples$}
\Comment{generate \# samples}

\State $mdv \gets 0.02$
\Comment{maximum divergence}
\State $G \gets$ \{ $<$ original dataset $>$ \} 
\Comment{ the class to augment}
\State $N \gets$ \{\}
\Loop  \space n\_samples \textbf{times}
\Comment{repeat for \# samples}

\State $a \gets$ uniform$(0,1)$  
\Comment{generate random number}
\State $b \gets a \times | G |$
\Comment{index of original sample selected}

\State $v \gets G[b]$
\Comment{value of original sample}

\State $coin \gets$ uniform$\{-1,1\}$
\Comment flip coin choice

\State $n \gets v \times mdv \times coin$
\Comment{generate value based on original}

\State $n \gets trunc (n)$
\Comment truncate value within range (0,1)
\State $N \gets N \cup \{n\}$
\Comment append to augmented set

\EndLoop

\EndProcedure

\end{algorithmic}
\end{algorithm}
For our setup, the $mdv$ value is set to 0.02 to generate samples with numerical values that deviate from the original by no more than 2\%. 
The chosen value was determined by trial and error to achieve the best results. 
It is worth noting that the large number of combinations generated by the augmentation algorithm minimizes the likelihood of overlapping samples. 
The probability of overlap is given by:
$p = \frac{1}{{(10 ^{x})}^y}$.
The significance value $x$ reflects the level of consideration given to the output of the first stage classifiers, whereas $y$ denotes the number of those incorporated in the scheme.
The significance relates to the number of numerical digits allotted for expressing the output scores, as specified in section \ref{DLStackEnsemble}.
In fact, in our setup, where $x=2$ and $y \in \{3,5\}$, this probability is indeed very low ($p = 10^{-10}$).
Eventually, with the application of the augmentation algorithm \ref{oversampling}, the data for training the second stage increased by over 43 times per run.
This resulted in an additional 80,000 samples being added to the 1,872 samples generated by the first stage for each subset.

\subsection{Experimental Setting}
\label{expSetting}
To maintain consistency with the results produced by state-of-the-art methods, such as \cite{Badjatiya:2017,Agwaral2019,Pitsilis2018} and others, we used the same 10-fold cross-validation scheme in our evaluation. 
The Deep Learning models used in both stages were implemented using Keras\footnote{https://github.com/fchollet/keras}.

In our experiment, we set the \emph{vocabulary size} for the first stage models based on uni-grams, bi-grams, and tri-grams to 25,000, 120,000, and 180,000, respectively, to match the size of the corpora and hence maintain compatibility with the encoded input vectors used in these models (see paragraph \ref{encoding}).
For the \emph{Deep Learning with Stacked Ensemble} scheme employed in the second stage, the vocabulary size must be equal to the size of the vector space of the combined first stage output, with value: $c \times 2 \times 10^{m}$.
As mentioned earlier, the value of $m$ indicates the level of weight given to the output of the first stage classifiers and denotes the number of digits used to express each two-tuple probability.
For instance, we set $m = 2$ to allow the \emph{Deep Learning Stacked Ensemble} classifier to recognize 100 different output states.
Indicated by the letter $c$ is the number of classifiers whose output is incorporated into the second stage.
We investigate two options:
i) $c=3$ which refers to the variation which considers just the word uni-grams,
and ii) $c=5$ which refers to the variation which considers the word bi-grams and tri-grams in addition to uni-grams.
For the above options, we set the \emph{vocabulary size} value for the second stage to 600 and 1,000 respectively.

\subsubsection{Model Training}
\label{main_model}

There is growing interest in using large \emph{batch size} in training DL models with \emph{stochastic gradient descent} since this can reduce learning time, \cite{Hoffer2017}. 
However, the use of a large \emph{batch size} is also known to improve the generalization performance.
The conventional \emph{stochastic gradient descent}, on the other hand, maintains a constant \emph{learning rate} throughout the training.
The general rule to counteract a large \emph{batch size} is to pick a suitable \emph{learning rate} or develop a good schedule for it.

Other algorithms, such as \cite{DeepLearningBook_Goodfellow_2016}, suggest using an adaptive \emph{learning rate}, in which the value of the \emph{learning rate} increases and decreases throughout the training process based on the error in the loss function. 
Despite that, no single solution exists so far to be the most effective for any given problem.
In our case, rather than adopting one of the most well-known adaptive learning policies, such as \emph{AdaGrad}, \emph{RMSProp}, or \emph{Adam}, we have chosen to create our pre-defined schedule with an adaptable \emph{learning rate} that will be adjusting as the training develops.
The general idea is that, while \emph{batch size} is set to a big fixed number, the \emph{learning rate} should be declining as the training progresses.
The rate of decline should depend on the impact on both the prediction accuracy and the error that occurred in the loss function approximation.

\begin{algorithm}
\caption{Model Training algorthim}
\label{training}
\begin{algorithmic}[1]
\Procedure{Train\_model}{} 

\State $max\_ep \gets 200$
\Comment{Max epochs to run}

\State $LR_{min} \gets 0.01$
\Comment{Low limit for learning rate}
\State $LR_{(ep=1)} \gets 0.05$
\Comment{Initial learning rate}

\State $ep2go \gets 40 $ 
\Comment{Set remaining epochs to run}
\State $i \gets 0$

\State $LR \gets LR_{(ep=1)}$
\Comment{Set learning rate}
\newline
\While{($ep2go > 0$) AND ($i < max\_ep$)} {
 
 \State $ep2go \gets (ep2go - 1)$
 \Comment{Decrease remaing epocs to run}
 \State \textbf{fit.model} with $LR$
 \State $i \gets (i + 1)$ 
 \State \textbf{compute} $error$    
 \Comment{Prediction Error}
 \State \textbf{compute} $accuracy$
 \Comment{Prediction Accuracy}
 \newline
 \If {BOTH $error$ \& $accuracy$ IMPROVED}
 
 \State $overfitting \gets (accuracy > training\_accuracy)$
 \Comment{Check overfitting}
 \If {NOT overfitting}
 \State \textbf{save.model}
 \State \textbf{extend} $ep2go$
 \Comment{Extend to run for more epochs}
 
 \EndIf
 
 \EndIf
 \newline
 \If {$LR > LR_{min}$} 
 \Comment{Decrease LR}
 \State{\textbf{update} $LR$}
 \EndIf
 
 \EndWhile}
\EndProcedure
\end{algorithmic}
\end{algorithm}

To train our models, we implemented an optimization rule described in algorithm \ref{training}.
In this algorithm, the value of the new \emph{learning rate} depends on the number of epochs ($ep$) elapsed since its last update, and is calculated using formula \ref{updateLR}.

\begin{equation}
\label{updateLR}
LR_{(e=ep)} = LR_{(ep=1)} \times (1 + dR * ep)^{-1} 
\end{equation}

$LR_{{(e=ep)}}$ denote as the new \emph{learning rate}, set at epoch $ep$, whereas $LR_{{(e=1)}}$ is the initial value given to the \emph{learning rate} in the first epoch.
For each model, the optimal value for the $LR_{{(e=1)}}$ was determined by trial and error. 
More particularly, for the training of the first stage OvR classifiers on uni-gram data, the $LR_{{(e=1)}}$ for the \emph{racism}, \emph{neutral}, and \emph{sexism} classifiers was set at 0.08, 0.08, and 0.10, respectively. 
As for the classifiers trained over bi-gram and tri-gram data for the \emph{racism} class, the $LR_{{(e=1)}}$ value was set to 0.05.
We applied the same trial and error policy for setting the decay rate ($dR$) value.
As such, its value was set to 0.15, 0.15, and 0.20 for the training of the unigram-based models for \emph{racism}, \emph{neutrality}, and \emph{sexism}, respectively.
Similarly, the chosen ($dR$) value for training the bi-gram and tri-gram models was 0.20 and 0.18, respectively.
Finally, the model training was allowed to run for a maximum of 200 epochs.
Concerning the \emph{batch size}, we set a different value for each stage.
More specifically, the DL classifiers in the first stage were trained with a \emph{batch size} of 1024, whereas the DL Stacked Ensemble classifier in the second stage was trained using a larger \emph{batch size} of 8192.
Both values remained constant throughout the training process.

In algorithm \ref{training}, the optimally trained state of the model, which corresponds to the best combination of Validation Accuracy and Error, is preserved throughout the training process to be used in model validation afterwards.
To achieve stability in the results produced, we repeated each experiment ten times, with the same data, and averaged the output values.

\subsection{Performance Metrics}

To demonstrate the effectiveness of our approach, we used standard metrics of classification accuracy, such as: \emph{Precision}, \emph{Recall}, and \emph{F-score}. 
More specifically, we report the \emph{micro averaged} values of the \emph{F-score} metric for two reasons.
Firstly, to maintain consistency with previous work in the field, and secondly, for the suitability of the \emph{micro average} for evaluating unbalanced test sets, as it incorporates class frequencies in the \emph{F-score} value calculation.
The \emph{Precision} (P) and \emph{Recall} (R) values for a specific class are derived from the True Positive (\emph{TP}), False Positive (\emph{FP}), and False Negative (\emph{FN)} instances by:
$P = \frac{TP}{(TP + FP)}$, $R = \frac{TP}{(TP + FN)}$.
\emph{P} is defined as the fraction of correctly classified tweets in a given class among all tweets classified in that class, whereas \emph{R} is the fraction of correctly classified tweets in a given class among all tweets in that class.
Furthermore, the \emph{F-score} metric is the harmonic mean of \emph{P} and \emph{R}, expressed as: $F = \frac{2 \cdot P \cdot R}{P + R}$.
In the case of multi-class classification, the total \emph{micro average} value of the \emph{F-score} is given by:
\begin{equation}
\label{eqn:microF}
F_{total} = \frac{\sum{F_{class-i}}}{\sum({n_{class-i}})}
\end{equation}
where $n_{class-i}$ denote as the population of a particular class $i$, whereas $F_{class-i}$ is the \emph{F-score} obtained for the tweets of that class. 

\subsection{Results - Discussion}

We present the most significant results from our experiments.
We contrast our results against other state-of-the-art algorithms for \emph{hate speech} detection, which have been evaluated over the dataset by \cite{Waseem2016b}.
In our comparison, we also include the algorithm developed by  \cite{Kapil:2020}, which was tested on the same dataset.
However, despite attaining the best \emph{F-score} ever, their reported results for this algorithm are not directly comparable to ours for two reasons: 
First, according to \cite{Kapil:2020}, the model training was done using a smaller subset, and second, the training process was reinforced by knowledge from other \emph{hate speech} datasets.

\begin{table*}[!ht]
   \centering
   \caption{Detailed Results for every Class Label}
   \label{tab:Detailed}
   
\begin{threeparttable}   
   \begin{tabular}{|c|c|c|c|c|c|c|c|c|c|}
   \hline
   \bf{Proposed} &       &   \multicolumn{3}{|c|}{\bf{3 OvR classifiers}}       & 
   \multicolumn{3}{|c|}{\bf{5 OvR classifiers}} & \multicolumn{2}{|c|}{\bf{Increase(\%)}\tnote{2}} \\ 
   \cline{3-10}
   \bf{Approach} & \bf{Class} & \bf{Precision} & \bf{Recall} & \bf{F-score}\tnote{1} & \bf{Precision} & \bf{Recall} & \bf{F-score}\tnote{1} & \bf{$F_R$} & \bf{$F_{tot}$} \\
   \hline
   \hline
           & \emph{Neutral} & 0.9537 & 0.9472 & 0.9504 & 0.9537 & 0.9485  & 0.9511 & &\\
   \multirow{1}{*}{Logistic} 
           & \emph{Racism}  & 0.7210 & 0.7474 & \bf{0.7330} & 0.7284 & 0.7464 & 0.7355 & 0.341 & 0.064 \\
   \multirow{1}{*}{Regression}
           & \emph{Sexism}  & 0.9960 & 0.9952 & 0.9956 & 0.9943 & 0.9950 & 0.9946 & & \\ 
   \hline
           & \emph{Neutral} & 0.9517 & 0.9470 & 0.9492 & 0.9524 & 0.9471 & 0.9497 & & \\
   \multirow{1}{*}{Random} 
           & \emph{Racism}  & 0.7158 & 0.7359 & 0.7246 & 0.7176 & 0.7404 & 0.7276 & 0.414 & 0.064 \\
   \multirow{1}{*}{Forest} 
           & \emph{Sexism}  & 0.9962 & 0.9947 & 0.9954 & 0.9960 & 0.9946 & 0.9953 & & \\    
   \hline
           & \emph{Neutral} & 0.9439 & 0.9544 & 0.9505 & 0.9475 & 0.9556 & 0.9515 & & \\
   \multirow{1}{*}{Gradient}
           & \emph{Racism} & 0.7396 & 0.7056 & 0.7198 & 0.7479 & 0.7101 & 0.7259 & 0.848 & 0.150 \\
   \multirow{1}{*}{Boosting} 
           & \emph{Sexism} & 0.9968 & 0.9949 & 0.9958 & 0.9967 & 0.9944 & 0.9956 & & \\    
   \hline
           & \emph{Neutral} & 0.9497 & 0.9497 & 0.9496 & 0.9492 & 0.9525 & 0.9507 & & \\
   \multirow{1}{*}{Ada} 
           & \emph{Racism}  & 0.7257 & 0.7270 & 0.7246 & 0.7360 & 0.7241 & 0.7282 & 0.497 & 0.129 \\
   \multirow{1}{*}{Boosting} 
           & \emph{Sexism}  & 0.9967 & 0.9924 & 0.9946 & 0.9968 & 0.9920 & 0.9944 & & \\    
   \hline
   
           & \emph{Neutral} & 0.9498 & 0.9530 & 0.9514 & 0.9544 & 0.9478 & 0.9510 & & \\
   \multirow{1}{*}{XG}
           & \emph{Racism}  & 0.7380 & 0.7228 & 0.7298 & 0.7234 & 0.7528 & 0.7372 & \bf{1.014} & 0.064 \\
   \multirow{1}{*}{Boosting}           
           & \emph{Sexism}  & 0.9957 & 0.9959 & 0.9958 & 0.9965 & 0.9944 & 0.9954 & & \\    
   
   \hline
           & \emph{Neutral} & 0.9533 & 0.9476 & 0.9487 & 0.9546 & 0.9477 & 0.9512 & & \\
   \multirow{1}{*}{Deep} 
           & \emph{Racism}  & 0.7218 & 0.7455 & 0.7328 & 0.7230 & 0.7537 & \bf{0.7375} & 0.6414 & 0.107 \\
   \multirow{1}{*}{Learning} 
           & \emph{Sexism}  & 0.9958 & 0.9947 & 0.9952 & 0.9961 & 0.9938 & 0.9950 & & \\
   \hline
   \multirow{1}{*}{Fixed} 
           & \emph{Neutral} & 0.9512 & 0.9515 & \bf{0.9514} & 0.9472 & 0.9598 & \bf{0.9535} & & \\
   \multirow{1}{*}{Criteria} 
           & \emph{Racism}  & 0.7332 & 0.7315 & 0.7323 & 0.7629 & 0.7049 & 0.7327 & 0.055 & \bf{0.172} \\
   \multirow{1}{*}{Ensemble}            
           & \emph{Sexism}  & 0.9958 & 0.9960 & \bf{0.9960} & 0.9955 & 0.9963 & \bf{0.9959} & & \\    
   
   \hline

   \end{tabular}
\end{threeparttable}

\begin{tablenotes}
\item[1] 1. The shown \emph{F,P} and \emph{R} are average values over many experiments.
\item[2] 2. The value regards the \emph{F-score} improvement for the \emph{Racism} class
alone.
\item[3] 3. The highlighted values indicate the top scores over all alternative approaches.
\end{tablenotes}

\end{table*}

\begin{table*}[!ht]
   \begin{adjustwidth}{-1in}{-1in}
   \centering
   \hspace*{-0.4in}
   \caption{Evaluation Results}
   \label{tab:Results}
   
   \begin{threeparttable}   
   \begin{tabular}{|c|c|c|c|c|}
   \hline
   \bf{2nd stage variation} & \bf{Characteristics} & \bf{Precision} & \bf{Recall} & \bf{F-Score\tnote{2}} \\
   \hline
   \hline
   Logistic Regression  &  \multirow{7}{*}{3 OvR classifiers}  & 0.9338   & 0.9324  & 0.9329  \\
   \cline{3-5}
   \cline{1-1}
   Random Forest  &   & 0.9318   & 0.9307 & 0.9311  \\
   \cline{3-5}
   \cline{1-1}
   Gradient Boosting &   & 0.9348  & 0.9322  & 0.9314 \\
   \cline{3-5}
   \cline{1-1}
   Ada Boosting  &    & 0.9318   & 0.9311  & 0.9312 \\
   \cline{3-5}
   \cline{1-1}
   XG Boosting   &    & 0.9332   & 0.9335  & 0.9333 \\
   \cline{3-5}
   \cline{1-1}
   Deep Learning  &    &  0.9335  & 0.9324  & 0.9329 \\
   \cline{3-5}
   \cline{1-1}
   Fixed Criteria Ensemble &  & 0.9336   & 0.9336  &  \bf{0.9336} \\
   \hline
   Logistic Regression  &  \multirow{7}{*}{5 OvR classifiers}  & 0.9344   & 0.9332  & 0.9335  \\
   \cline{3-5}
   \cline{1-1}
   Random Forest  &   & 0.9325   & 0.9314 & 0.9317  \\
   \cline{3-5}
   \cline{1-1}
   Gradient Boosting &   & 0.9330   & 0.9326  & 0.9328 \\
   \cline{3-5}
   \cline{1-1}
   Ada Boosting  &    & 0.9327   & 0.9326  & 0.9326 \\
   \cline{3-5}
   \cline{1-1}
   XG Boosting   &    & 0.9347   & 0.9333  & 0.9338 \\
   \cline{3-5}
   \cline{1-1}
   Deep Learning  &    &  0.9347  & 0.9333  & 0.9339 \\
   \cline{3-5}
   \cline{1-1}
   Fixed Criteria Ensemble &  & 0.9344   & 0.9361  &  \bf{0.9351} \\
   \hline
   \hline
   \cite{Waseem2016b} & Logistic Regression & 0.7293 & 0.7774 & 0.7393 \\
                      &       & & & \\
   \hline
   \cite{Waseem2016} & Logistic Regression  & 0.9159 & 0.9292 & 0.9153 \\
                     &   (Extended dataset)    & & & \\
   \hline
   \cite{Park2017}  & 2 stage HybridCNN           & 0.8270 & 0.8270 & 0.8270 \\ 
                    &    (Word Vect. / Char Vect.)  &       &        & \\
   \hline
   \cite{Pitsilis2018} & Enssemble of LSTM & 0.9305 & 0.9334 & 0.9320 \\
                           & + user history & & & \\
   \hline
   \cite{mishra-etal-2018} & Graph-based user profiling & 0.8757 & 0.8766 & 0.8757 \\
                           & & & & \\
   \hline
   \cite{mishra-etal-2018-neural} & GRU-based RNN & 0.8128 & 0.7784 & 0.7937 \\
                           & & & & \\
   \hline
   \cite{bodapati-etal-2019-neural} & BERT-based & N/A\tnote{1} & N/A\tnote{1} & 0.8290 \\
                           & (used reduced dataset) & & & \\
   \hline
   \cite{meyer-gamback-2019-platform} & CNN - RNN - LSTM & 0.8414 & 0.8414 & 0.8414 \\
                           &  & & & \\
   \hline
   \cite{swamy-etal-2019-studying} &  BERT & N/A\tnote{1} & N/A\tnote{1} & 0.5837 \\
                           &  & & & \\
   \hline
   \cite{Agwaral2019}\tnote{3} & BiLSTM with Attention & 0.8470 & 0.8410 & 0.8430   \\
                           & (amended scores) & & & \\
   \hline
   \cite{Badjatiya:2017}\tnote{3} & LSTM + Random Embedding & 0.8230 & 0.8210 & 0.8070 \\
                         & + GBDT (amended scores)                 &        &        &        \\
   \hline
   \cite{Kapil:2020}
   & Multi-task LSTM  & N/A\tnote{1} & N/A\tnote{1}  &  \underline{0.9410} \\
   &  (trained on mixed datasets) &        &        &        \\
   \hline
   
   \end{tabular}
   \end{threeparttable}
   
\begin{tablenotes}
\item[1] 1. A score marked as N / A indicates an unavailable result from the original paper
\item[2] 2. The \emph{F, P}, and \emph{R} values shown are averages of multiple experiments
\item[3] 3. Used amended scores by \cite{Arango:2019}
\end{tablenotes}
   
   \end{adjustwidth}
\end{table*}

From the results presented in tables \ref{tab:Detailed} and \ref{tab:Results}, we observe the following:

\begin{itemize}
    \item Our approach outperforms all the alternative algorithms by other researchers in the field comprised in our study. 
    That includes both the supervised and the unsupervised learning solutions.
    More specifically, among the current state-of-the-art, our best performing model achieved an \emph{F-score} of 0.9351, compared to 0.9320 by the second-best solution by \cite{Pitsilis2018}.
    \item In terms of which of the proposed alternative algorithms for the second stage task performs best, we observe that the highest \emph{F-score} = 0.9351 was attained by the \emph{Fixed Criteria Ensemble} scheme, while the \emph{Deep Learning} scheme achieved the second-best score with 0.9339.
    We attribute the reason for the inferior performance of the \emph{Deep Learning} scheme to the limited amount of data used for training.
    In particular, training the first stage classifier over a small dataset of only 16k tweets, as we did, produced insufficient data for training the second stage classifier.
    Testing on a larger dataset of labelled tweets, on the other hand, could be in favour of the \emph{Deep Learning} scheme.
    \item Regarding the potential benefit in \emph{racism} detection provided by incorporating bi-grams and tri-grams features into the training process, we note that there is indeed a performance benefit (see column labelled "5 \emph{OvR} classifiers" in Tables \ref{tab:Detailed} and \ref{tab:Results}) over using only uni-gram features (see column labelled "3 \emph{OvR} classifiers" in Tables \ref{tab:Detailed} and \ref{tab:Results}).
    More specifically, as can be seen in Table \ref{tab:Detailed} (see column labelled "$F_R$", showing the actual improvement in terms of the \emph{F-score} for the \emph{racism} class), in all schemes we introduced, \emph{racism} detection alone was benefited from the use of additional features by at least 0.055\%.
    As can also be observed, \emph{XG Boosting} is the most affected algorithm for the second stage, raising the \emph{$F_R$} by 1.014\%, followed by \emph{Gradient Boosting}, which improves the \emph{$F_R$} by only 0.848\%.
    On the contrary, the \emph{Fixed Criteria Ensemble} is the least affected alternative for detecting \emph{racism}.
    For the remaining classes of \emph{sexism} and \emph{neutrality}, the incorporation of extra features resulted in a small benefit observed.
    Regarding the overall \emph{F-score} for all three classes (see the column labelled "$F_{tot}$" in table \ref{tab:Detailed}), the \emph{Fixed Criteria Ensemble} appears to be the most benefitted scheme by the incorporation of bi-gram and tri-gram features, by 0.172\%.
    
    \item Compared to other state-of-the-art algorithms, the use of pre-trained word embeddings in our model provides a noticeable benefit in terms of classification accuracy.
    From the results, we observe that our technique outperforms all other algorithms examined on the same dataset, such as \cite{Park2017,mishra-etal-2018,bodapati-etal-2019-neural,meyer-gamback-2019-platform,swamy-etal-2019-studying,Agwaral2019,Badjatiya:2017}, that incorporate various types of embeddings into the input data encoding.
    Considering the above finding, we conclude that the use of word embeddings is not a panacea for improving performance.
    
    In addition, as opposed to other single-stage solutions that do not use pre-trained word embeddings, our proposed scheme of OvR Deep Learning classifiers still does better in the multi-class \emph{hate speech} detection task.
    For instance, our approach outperformed \cite{Waseem2016b,Waseem2016} LR-based models by 26.49\% and 2.16\%, respectively, and it is also 0.34\% more efficient than the single-stage DL-based solution by \cite{Pitsilis2018}.
 
\end{itemize}

Given the above observations on the results, one of the future challenges may be to evaluate our supervised learning methods on datasets large enough to produce adequate data to train a second stage classifier.
Since our model outperforms similar approaches, like \cite{Waseem2016b,Pitsilis2018}, without necessitating any user profile information to incorporate as a feature in the training process, it is clear that high-quality \emph{hate speech} detection is still viable without compromising user privacy. 
Despite failing to outperform the \emph{Fixed Criteria Ensemble}, the \emph{DL with Stacked Ensemble} scheme remains, in our opinion, the most promising solution for the second stage classifier (see section \ref{DLStackEnsemble}). 
The main reason for this is that, unlike the \emph{Fixed Criteria Ensemble}, the \emph{Stacked Ensemble} is well suited to work in a \emph{Transfer Learning} scheme.
It may, for example, be modified to support complex scenarios like combining several sources of labelled data into a single vector and then using that to train the core classifier. 
However, more testing is required to optimize a number of parameters, such as the number of states to discriminate from the output of the first stage, which will be utilized to train the \emph{Stacked Ensemble}.

Finally, among the six alternative schemes that we proposed and tested as second stage classifier options, we recommend the following, as the most promising: 
\emph{i)} the \emph{Fixed Criteria Ensemble}, for its simplicity and high efficiency in determining the class output, 
\emph{ii)} the \emph{Deep Learning with Stacked Ensemble}, for its applicability to \emph{Transfer Learning} setups for classifying various forms of \emph{hate speech}, beyond \emph{sexism} and \emph{racism}.

\subsection{Generalization}
There has recently been a trend toward using automatic detection methods that work without modification across a wide range of online platforms and areas.
A technique to achieve this for natural language processing tasks, such as \emph{hate speech} classification, is to adapt \emph{Complete Transfer Learning}, a type of \emph{Transfer Learning}, in the model training.
In this manner, without having to retrain a model, the knowledge gained from training on one dataset may be used to improve \emph{hate speech} detection performance on new data.

\emph{Cross-Dataset Training and Testing} (a.k.a Generalization Testing) is a method of determining how effectively a system can achieve the above goal by testing in a real-world scenario.
The experiments reported in this section aim to investigate how effectively our model generalizes to other datasets from the same domain after being trained on the \cite{Waseem2016b}, in contrast to other techniques.

For our comparative study, we selected two state-of-the-art algorithms developed by \cite{Badjatiya:2017} and \cite{Agwaral2019}
which
\cite{Arango:2019} has documented in a generalization study for \emph{hate speech} detection.
Another algorithm by \cite{Pitsilis2018}, which performed well on the same dataset, is also included in our comparison.
Overall, we tested two different variations of that algorithm\footnote{Source code found at: https://github.com/gpitsilis/hate-speech}. 
The first variation, which has achieved the best \emph{F-score}=0.9320 (see table \ref{tab:Results}), employs additional features related to users' tendency toward hatred behaviour in tweeting, whereas the second variation does not.

As far as the datasets used in the comparative generalization study, while all models were trained on \cite{Waseem2016b} data, for test set we chose a publicly available dataset by \cite{SemEval2019} as the most suitable.
This dataset was made available via a shared task for offensive language detection, known as \emph{OffensEval 2019: Identifying and Categorizing Offensive Language in Social Media} (a.k.a \emph{SemEval-2019}), and we chose it because of its widespread use in benchmarking numerous models.
The original \emph{SemEval} dataset contains over 9,000 text samples, each with a pair of binary labels. 
While the first label shows whether toxic language is present in a given sample, the second label indicates whether or not that toxic language is aggressive.

\subsubsection{Model adjustments}
\label{adaptations}

Due to label discrepancies between the training and testing datasets we used for the generalization study, we made several changes in the evaluation task.
Although the labelling in both the training and testing data sets is suited for 3-way classification, drawing any conclusions about the test result would not be feasible due to the contextually different annotations used in these sets.
To overcome this, we made appropriate changes to the evaluation task, such as rearranging the labels in the test data and transforming the classifier model schemes to fit the testing scenario. 
Concerning the test data, the aim was to take into account only those series of class labels that share the same context between the training and test sets.
To achieve this, we focused exclusively on the primary label of the test set, which indicates the presence of \emph{hate speech} in the sample, disregarding the secondary one, which further differentiates \emph{hate speech} into aggressive and non-aggressive forms.
Based on this, the 9k sample set was divided into 3783 hateful and 5217 non-hateful samples. 

In terms of the transformations needed in the trained model schemes, our key objective was to fit the existing OvR classifiers to the available test data such that the resulting classifier would produce normalized output in the form of Hate/No-Hate. 
We devised three different testing scenarios in total, which are detailed below:

\begin{enumerate}[label=(\roman*)]

    \item Single classifier model. 
    This model only employs the trained Neutral/Non-neutral classifier ($H_N$), which can distinguish Hatred from Non-Hatred text content in input $x$, as follows:
    
    $
    H_{class}(x) = 
    \begin{cases}
    H    : p_N(x) < 0.5 \\
    \neg H : p_N(x) \ge 0.5 \\
    \end{cases}
    $
    
    Based on this reasoning, any input sample $x$ from the test set classified as ’Non-neutral’, will be considered \emph{hate speech}
    ($H_{class}(x) = H$).
    
    \item Three classifiers model ($H_N$, $H_S$, $H_R$). 
    This method utilizes the main OvR classifiers trained on uni-gram input.
    Eqn.\ref{E1} is modified to generate binary output as follows:
    
    $
    H_{class}(x) = 
    \begin{cases}
    H    : p_N(x) < max[p_R(x), p_S(x)] \\
    \neg H : p_N(x) \ge max[p_R(x) ,p_S(x)] \\
    \end{cases}
    $

    $p_{(i)}(x)$ denotes the positive class output probability of classifier $i$ for input text sample $x$, with $i \in \{l_N,l_S,l_R\}$.
    The class label determined for sample $x$ can be either:
    ($H_{class}(x) = H$) if there is \emph{hate speech} in $x$, or ($H_{class}(x) = \neg H$) otherwise.

    \item Five classifiers model ($H_N$, $H_S$, $H_R$, $H_{R_2}$, $H_{R_3}$). 
    This one is an extension of the three classifiers model, which combines the output of trained word unigram-based classifiers ($H_N$, $H_S$, $H_R$) with the bi-gram and tri-gram based ($H_{R_2}$, $H_{R_3}$), that served the purpose of \emph{racism} detection.
    As such, the class label $H_{class} \in \{H, \neg H\}$ determined for sample $x$ is given as follows:
    
    $
    H_{class}(x) = 
    \begin{cases}
    H    : p_N(x) < max[p_S(x), p_R(x), p_{R_2}(x), p_{R_3}(x)] \\
    \neg H : p_N(x) \ge max[p_S(x) ,p_R(x), p_{R_2}(x), p_{R_3}(x)] \\
    \end{cases}
    $

    $p_{R_2}(x)$ and $p_{R_3}(x)$ denote the positive class output probabilities for input text sample $x$, of word bi-grams and tri-grams classifiers, respectively, trained on \emph{racism} detection.
    
\end{enumerate}

\subsubsection{Data Pre-processing}

The differences in terminology, idioms, and message length between the training and test data necessitated the application of filtering on the test data before the validation task.
Filtering on datasets is a common pre-processing task adopted by many researchers in the area of text classification, such as,
\cite{Kapil:2020,meyer-gamback-2019-platform,nikhil-etal-2018,zhang2019identifying,mishra-etal-2018-neural}.
The main objective of the pre-processing in our particular case was to remove the noisy text while preserving any key features. 
Our filtering scheme consists of seven steps applied to each test tweet in a specific order, as shown in Table \ref{tab:filtering}.

\begin{table*}[htp]
\centering
\caption{Filtering scheme applied on \emph{SemEval-19} test set}
\label{tab:filtering}
   
\begin{tabular}{|c|l|}
\hline
\emph{Order} & \emph{Operation} \\
\hline
\hline
1. & Removing any quote (") characters from the text \\ 
\hline
2. & Removing
usernames from the text ( character strings beginning with @ ) \\
\hline
3. & Removing any URLs ( character strings beginning with http:// or https:// ) \\
\hline
4. & Removing any punctuation symbols, such as comma (,) colon (:) and semicolon (;) \\
\hline
5. & Deleting the hash (\#) symbol prefix from any hash-tags in the text \\
\hline        
6. & Removing any retweet symbols (RT) from the text \\
\hline
7. & Removing non-standard ASCII characters \\
\hline
\end{tabular}

\end{table*}

For highlighting the impact of filtering, our results also include the case of not applying filtering to the plain test set.

\subsubsection{Setup - Metrics}

Among the alternative schemes proposed in this paper, we chose to evaluate the \emph{Fixed Criteria Ensemble} only, which we describe in section \ref{intuitive}.
This one was chosen not only for its simplicity but also because it performs the best in terms of \emph{F-score} rating.

To maintain consistency with the main experiment, we implemented the generalization task as part of the main 10-fold cross-validation scheme described in section \ref{expSetting}.
Therefore, each trained model out of the 10 created in the main cross-validation experiment was used unmodified for validation across the entire test set.
The reported performance score indicates the average over 10 folds, repeated for 10 experiment runs.
To quantify the performance in the generalisation experiment, we also used the metrics of \emph{Precision} and \emph{Recall}, with the reported \emph{F-score}.
Finally, the formula for the micro-average computation of the total \emph{F-score} (see eqn.\ref{eqn:microF}) was modified for two-class output (Hatred / Non-Hated), according to the requirements outlined in section \ref{adaptations}.

\subsubsection{Generalization Results - Discussion}

In table \ref{tab:Generalization} we show the performance figures.
For the methods by \cite{Badjatiya:2017} and \cite{Agwaral2019}, the presented \emph{F-score} values in the table are those reported by \cite{Arango:2019}.
Also, in the same table, the reported values of the method by \cite{Pitsilis2018} emerged by reproducing their algorithm over the filtered and the unfiltered test set.
We tested two variations of this method: i) one that uses a single classifier and ignores the user's posting history, and ii) an ensemble scheme, which the authors mention as the best performing of those tested. 
The latter referred to as (O+NS+RS+NR+NRS) in their paper, consists of five classifiers, each of which combines different user-related features.
Each feature is associated with a user's tendency to express \emph{sexism} (S), \emph{racism} (R), or \emph{neutrality} (N) in their posts.
\begin{table*}
   \centering
   \caption{Results of generalization to \emph{SemEval-2019} dataset}
   \label{tab:Generalization}   
   \begin{tabular}{|c|c|c|c|c|}
   \hline
   \bf{Approach -} & \bf{Variation} &     
   \multicolumn{3}{|c|}{\bf{F-Score}}   \\
   \cline{3-5}
\bf{Evaluation data}     &  \bf{(trained classifiers used)}     &    \bf{Non-Hatred} & \bf{Hatred} & \bf{Total}  \\ 
   \hline
   \hline
    \multirow{2}{*}{Proposed model on} 
           & \multicolumn{1}{l|}{(i) \emph{$H_N$}} & 0.1150 & 0.5747 & 0.3082 \\
        \multirow{2}{*}{ \emph{plain} test-set } & \multicolumn{1}{l|}{(ii) \emph{$H_N,H_R,H_S$}} & 0.1288 & 0.5677 & 0.3133 \\
           & \multicolumn{1}{l|}{(iii) \emph{$H_N,H_R,H_S$} \& \emph{$H_{R_2}, H_{R_3}$}} & 0.1177 & 0.5723 & 0.3088 \\ 
   \hline
     \multirow{2}{*}{Proposed model on} 
            & \multicolumn{1}{l|}{(i) \emph{$H_N$}} & 0.5160 & 0.5694 & 0.5384 \\
       \multirow{2}{*}{ \emph{filtered} test-set }    & \multicolumn{1}{l|}{(ii) \emph{$H_N,H_R,H_S$}} & 0.5541 & 0.5421 & \underline{0.5491} \\
           & \multicolumn{1}{l|}{(iii) \emph{$H_N,H_R,H_S$} \& \emph{$H_{R_2}, H_{R_3}$}} & 0.5468 & 0.5494 & 0.5479 \\ 
   \hline
   \hline
           \multicolumn{2}{|c|}
           {Method by \cite{Badjatiya:2017}}
           & 0.7390 & 0.2110 & 0.5106 \\
   \hline
           \multicolumn{2}{|c|}
           {Method by \cite{Agwaral2019}}
           & 0.7290 & 0.2160 & \bf{0.5970} \\
   \hline

   \multirow{1}{*}{Method by \cite{Pitsilis2018}} 
        & \multicolumn{1}{l|}{(a) user features not considered} & 0.2482 & 0.5754 & 0.3858 \\
   \multirow{1}{*}{ on \emph{plain} test-set }    & \multicolumn{1}{l|}{(b) user-history features considered} & 0.2114 & 0.5273 & 0.3442 \\
   \hline
     
   \multirow{1}{*}{Method by \cite{Pitsilis2018}} 
        & \multicolumn{1}{l|}{(a) user features not considered} & 0.5670 & 0.5076 & 0.5420 \\
   \multirow{1}{*}{ on \emph{filtered} test-set }    & \multicolumn{1}{l|}{(b) user-history features considered} & 0.6188 & 0.4434 & 0.5451 \\
   \hline
     
   \end{tabular}
\end{table*}
We make the following observations based on the results shown in table \ref{tab:Generalization}:

\begin{itemize}
    \item 
    As shown in Table \ref{tab:Generalization}, despite being outperformed by the \cite{Agwaral2019} method, our approach ultimately achieves a higher F-score than \cite{Badjatiya:2017} and \cite{Pitsilis2018}.
    Moreover, despite outperforming only two out of the three other state-of-the-art algorithms (\emph{F} = 0.5491), our approach still obtains a good balance between the F-scores for the \emph{Hatred} and the \emph{No-Hatred} classes, with values of 0.5421 and 0.5541 respectively, something that the other algorithms fail to achieve.
    \item As also observed, the alternative schemes of our technique that used trained classifiers with extended features for the \emph{racism} class (see trained classifiers $H_{R_2}$, $H_{R_3}$ in Table \ref{tab:Generalization}) do not seem to provide a significant performance benefit when tested on unseen data. 
    On the contrary, as mentioned earlier (see table \ref{tab:Detailed}), there is such a benefit when tested on the \cite{Waseem2016b} dataset.
    We can attribute this to the necessary adjustments made to the evaluation task to meet the needs of the generalization study.
    As a consequence, while the trained model could only deal with the \emph{hate speech} categories of \emph{racism} and \emph{sexism}, any information communicated by word bi-grams and tri-grams would remain largely unrelated between the two datasets. 

    \item Filtering the test data prior to classification has a positive effect on performance when compared to the classification of the text in its plain form.
    As we observed, certain symbols and special characters, such as '\#' for hashtags, 'RT' for retweets, or '@' for indicating usernames, which we filtered out of the test set, are mainly encountered in \emph{sexist} tweets in \cite{Waseem2016b} training dataset.
    For instance, of the 3160 \emph{sexism} labeled tweets, we found 293 URLs, 1367 'RT' symbols, and 2511 '@' characters.
    The effect of filtering on performance may be due to the bias of trained models in classifying any input message containing these symbols as offensive.
    The findings suggest that filtering may help reduce bias and thus improve classification accuracy.
    \item All proposed alternatives of our approach based on a single classifier, using only the Hatred / No-Harted class (cases denoted by \emph{i} in Table \ref{tab:Generalization}), fail to outperform any alternative that incorporates multiple classifiers (cases denoted by \emph{ii} \& \emph{iii} in Table \ref{tab:Generalization}).
    Instead, the latter seems to have a marginal advantage of 1.36 \% on average.
    In classification problems, it is usual for ensemble-based techniques to obtain greater classification accuracy than single classifiers, and we believe our particular case falls under this category.
    
    \item When compared to the second-best alternative by \cite{Pitsilis2018}, which we also included in the main experiment (see Table \ref{tab:Results}), we observe that when the filtered test set is used, our proposed method (variation \emph{ii} in Table \ref{tab:Generalization} with \emph{F-score} 0.5491) has a marginal advantage over the above algorithm when tested on unseen data (cases denoted by the letters \emph{a},\emph{b} in Table \ref{tab:Generalization} with \emph{F-score} 0.5420 and 0.5451).
    Furthermore, even though our approach does not employ user history-related features, it still generalizes better by 0.734 \% than the variation by \cite{Pitsilis2018} that do (case denoted by the letter \emph{b} in Table \ref{tab:Generalization}).
    
\end{itemize}

Despite outperforming the existing methods and generalising better than the current state-of-the-art, we believe our proposed algorithm's full potential has yet to be proven.

We find three plausible explanations for our algorithm's poor cross-dataset performance in the generalization experiment.
First, the necessary modifications made to the evaluation scheme as a result of the different labelling used in the training and test data is a likely reason for the reduced score.
Second, there is a strong user bias in the training dataset, as reported by other researchers in the field, \cite{Arango:2019}, which may have seriously affected the measured performance in the main experiment. 
The lack of such bias in the generalization dataset might explain the performance drop observed.
Third, because we encoded the training data using a word-frequency vectorization strategy, many words in the test set were not indexed and so remained unknown to the classifier.
For instance, only around half of the 18367 unique words from the test set exist in the training set.

In spite of the low score obtained when tested on unseen data, our algorithm performs satisfactorily well compared to other algorithms.
To give an example of a typical performance reduction that may occur in generalization tests, we mention a case of an algorithm reported by \cite{Arango:2019}, which, after being trained on the \cite{Waseem2016b} dataset, exhibited a dramatic drop in the value of the \emph{F-score} to 0.1 when evaluated on Wikipedia data.

\section{Conclusions and future work}
\label{conclusion}

Automated detection of abusive language in on-line media has become a big challenge in the recent years.
In this paper, we proposed a novel approach that is an extension of an existing algorithm that uses LSTM neural networks to address the above challenge. 
The proposed two-stage scheme, designed for multi-label data, employs \emph{One-vs-Rest} classifiers and is properly tuned and adjusted to determine the type of harassment in tweets, with high accuracy.
The encoding of the input text, in a form independent of pre-trained word embeddings, is one of the innovations of the proposed approach.
This feature translates into the ability to capture idioms and slang expressions, that may likely exist in the text.
In addition, the model only utilizes the textual content of tweets to classify a new posting without requiring knowledge of users' tweeting history, thus respecting their privacy.

This paper has made several contributions which can be summarized as follows:
\emph{i)} A model for fine-grained abusive text detection with language-agnostic capability, suitable for capturing \emph{hate speech} indirectly expressed through obfuscated insulting terms.
\emph{ii)} A comparative experimental evaluation against other state-of-the-art solutions over a publicly available corpus of 16k labelled tweets.
\emph{iii)} A generalization study by testing on another 9k tweet dataset and comparison with state-of-the-art algorithms.
\emph{iv)} An implementation built on top of Keras toolkit.

The experimental results showed the superiority of our proposed design over the existing state-of-the-art algorithms in distinguishing the two types of \emph{hate speech} we tested, \emph{racism} and \emph{sexism}, from regular text.
Also, under certain conditions, the proposed model generalizes well to other datasets.
Most important, despite not using user history information, the model generalizes better than the current state-of-the-art.
Our experimentation has also revealed that fine-tuning is necessary for Deep Learning-based models to improve classification performance.
Furthermore, the experimental results support the initial hypothesis that improving the \emph{hate speech} classification performance does not necessitate a solution based on linguistic rules and vocabularies.
Due to the algorithm's inherent suitability for language-agnostic solutions, we intend to extend our investigation to texts written in other languages.

Lastly, we believe that experimenting with a single data set, as we did, has little to offer in terms of improving classification performance.
Integrating multiple datasets into a \emph{Transfer Learning} setting, on the other hand, appears to be a quite promising future direction for the proposed design. 
For NLP classification tasks like the one we looked at in this article, it is vital to be able to explain convincingly the decisions made by an algorithm, especially when it comes to consumer services, \cite{risch-krestel-2018-delete}.
In the context of freedom of speech, decision justification is critical in comment moderation so that the classification result is convincing that it is not the result of censorship.
Therefore, another potential area of future research could be the explainability of \emph{hate speech} classification.

The code and results associated with this paper will be available on-line, soon at:\\ \url{https://github.com/gpitsilis/} and \\ \url{https://github.com/gpitsilis/Generalization-test-hate-speech}

\biboptions{authoryear}

\bibliography{bibliography.bib}

\end{document}